\pdfoutput=1
\documentclass[letterpaper,twocolumn,fleqn]{article} 

\usepackage{ist}

\usepackage{graphicx}
\usepackage{epstopdf}
\usepackage{amsmath,amsfonts}
\usepackage{subfig}
\usepackage{multirow}
\usepackage{color}
\usepackage{comment}

\DeclareGraphicsExtensions{.eps,.pdf}
\graphicspath{{./figures/}}
\DeclareMathAlphabet{\mathcal}{OMS}{cmsy}{b}{n}

\title{Tubule Segmentation of Fluorescence Microscopy \\ Images Based on Convolutional Neural Networks \\ With Inhomogeneity Correction}
\author{Soonam Lee $^a$, Chichen Fu $^a$, Paul Salama $^b$, Kenneth W. Dunn $^c$, and Edward J. Delp $^a$ \\
$^a$Video and Image Processing Laboratory (VIPER), School of Electrical and Computer Engineering, Purdue University, West Lafayette, Indiana 47907, USA \\
$^b$Department of Electrical and Computer Engineering, Indiana University, Indianapolis, Indiana 46202, USA\\
$^c$Division of Nephrology, School of Medicine, Indiana University, Indianapolis, Indiana 46202, USA\\
}

\date{} 

\begin{document}

\maketitle

\thispagestyle{empty} 

\begin{abstract}
Fluorescence microscopy has become a widely used tool for studying various biological structures of in vivo tissue or cells. However, quantitative analysis of these biological structures remains a challenge due to their complexity which is exacerbated by distortions caused by lens aberrations and light scattering. Moreover, manual quantification of such image volumes is an intractable and error-prone process, making the need for automated image analysis methods crucial. This paper describes a segmentation method for tubular structures in fluorescence microscopy images using convolutional neural networks with data augmentation and inhomogeneity correction. The segmentation results of the proposed method are visually and numerically compared with other microscopy segmentation methods. Experimental results indicate that the proposed method has better performance with correctly segmenting and identifying multiple tubular structures compared to other methods.
\end{abstract}

\section{Introduction}
\label{sec:intro}
Recent advances in fluorescence microscopy imaging, especially two-photon microscopy, have enabled the imaging of cellular and subcellular structures of living tissue \cite{bib:BHP2008, bib:Helmchen2005, bib:Dunn2002}. This has resulted in the generation of large datasets of 3D microscopy image volumes, which in turn need automatic image segmentation techniques for quantification. However, the quantitative analyses of these datasets still pose a challenge due to light scattering, distortion created by lens aberrations in different directions, and the complexity of biological structures \cite{bib:Murphy2012}. The end result is blurry  image volumes with poor edge details that become worse in deeper tissue depths.

There have been various techniques developed for segmentation of microscopy images. One widely used class of methods is based on the active contours technique which minimizes an energy functional to fit contours to objects of interest \cite{bib:Kass1988, bib:Delgado-Gonzalo2015}. Early version of active contours \cite{bib:Kass1988} generally produced poor segmentation results since the segmentation results are noise sensitive and initial contour dependent. An external energy term which convolves a controllable vector field kernel with an image edge map was presented in \cite{bib:BLi2007} to address the noise sensitive problem. Similarly, the Poisson inverse gradient was introduced to determine initial contours locations to segment microscopy images in \cite{bib:BLi2008}. Moreover, active contours methodology has been integrated with a region-based segmentation method that poses the segmentation problem as an energy equilibrium problem between foreground and background regions \cite{bib:Chan2001}. In addition, this region-based active contours technique was extended to fully utilize 3D information to identify foreground and background voxels \cite{bib:Lorenz2013}. More recently, this 3D region-based active contours was combined with 3D inhomogeneity correction to provide better segmentation since this technique takes into consideration inhomogeneities in volume intensity \cite{bib:SLee2017}. Additionally, a new segmentation method known as Squassh that couples image restoration and segmentation using a generalized linear model and Bergman divergence was introduced \cite{bib:Paul2013}, whereas a method that combined with detecting primitives based on nuclei boundaries and identifying nuclei region using region growing was demonstrated in \cite{bib:Arslan2013}. Alternatively, combination of multiresolution, multiscale, and region growing methods using random seeds to perform multidimensional segmentation was described in \cite{bib:Srinivasa2009}.  

As indicated above, florescence image segmentation still remains a challenge problem. Tubule, a biological structure with a tubular shape, segmentation is even more challenging since tubular shape and orientation is varied without known patterns. Also, since typical tubular structures have hollow shapes with unclear boundaries, traditional energy minimization based methods such as active contours have failed segmenting tubular structures \cite{bib:SLee2015}. 
There has been some work particularly focusing on tubular structure segmentation. A minimal path based approach was described in \cite{bib:Benmansour2011, bib:HLi2007} where tubule shape is modeled as the envelope of a family of spheres (3D) or disk (2D). Similarly, a new approach for 3D human vessels segmentation and quantification using 3D cylindrical parametric intensity model was demonstrated in \cite{bib:Worz2004}. Also, multiple tubule segmentation technique that combined with level set methods and the geodesic distance transform was introduced in \cite{bib:Fakhrzadeh2012}. More recently, one method used to segment tubular structures was delineating tubule boundaries followed by ellipse fitting to close the boundaries while considering intensity inhomogeneity \cite{bib:SLee2015}. Another method known as Jelly filling \cite{bib:Gadgil2016b} utilized adaptive thresholding, component analysis, and 3D consistency to achieve segmentation, whereas a method for tubule boundary segmentation used steerable filters to generate potential seeds from which to grow tubule boundaries followed by tubule/lumen separation and 3D propagation to generate segmented tubules in 3D \cite{bib:DHo2017a}. Previous methods, however, focused on segmenting boundaries of tubule membrane. Since some tubule membranes are not clearly delineated in fluorescence microscopy image volume, finding tubule boundaries may not always result in identifying individual tubule regions.

Convolutional neural network (CNN) has been used to address segmentation problems in biomedical imaging \cite{bib:Litjens2017}. The fully convolutional network \cite{bib:FCN} introduced an encoder-decoder architecture for semantic segmentation. U-Net \cite{bib:UNet} is a 2D CNN based method utilizing this encoder-decoder architecture with connecting intermediate stages of downsampling and upsampling to preserve information. U-Net can be used segment complex biological structure in microscopy images. Similarly, in \cite{bib:DCAN} a U-Net trained on cell objects and contours was used to identify tubular structures. Additionally, a multiple input and multiple output structure based on a CNN for cell segmentation in fluorescence microscopy images was demonstrated in \cite{bib:MIMONet}. Also, a nuclei segmentation method that combined with a 2D CNN and a 3D refinement process was introduced in \cite{bib:CFu2017}.

In this paper, we present a method for segmenting and identifying individual tubular structure based on a combination of intensity inhomogeneity correction, data augmentation, followed by a CNN architecture. Our proposed method is evaluated at object-level metrics as well as pixel-level metrics using manually annotated groundtruth images of real fluorescence microscopy data. Our datasets are comprised of images of a rat kidney labeled with a phalloidin which labels filamentous actin collected using two-photon microscopy. A typical dataset we use in our studies consists of two tissue structures, the base membrane of the tubular structures and the brush border which is generally located interior to proximal tubules. Our goal here is to segment individual tubules enclosed by their membranes. 


\section{Proposed Method}
\label{sec:method}

\begin{figure}[htbp!]
	\centering
	\includegraphics[width=0.5\textwidth]{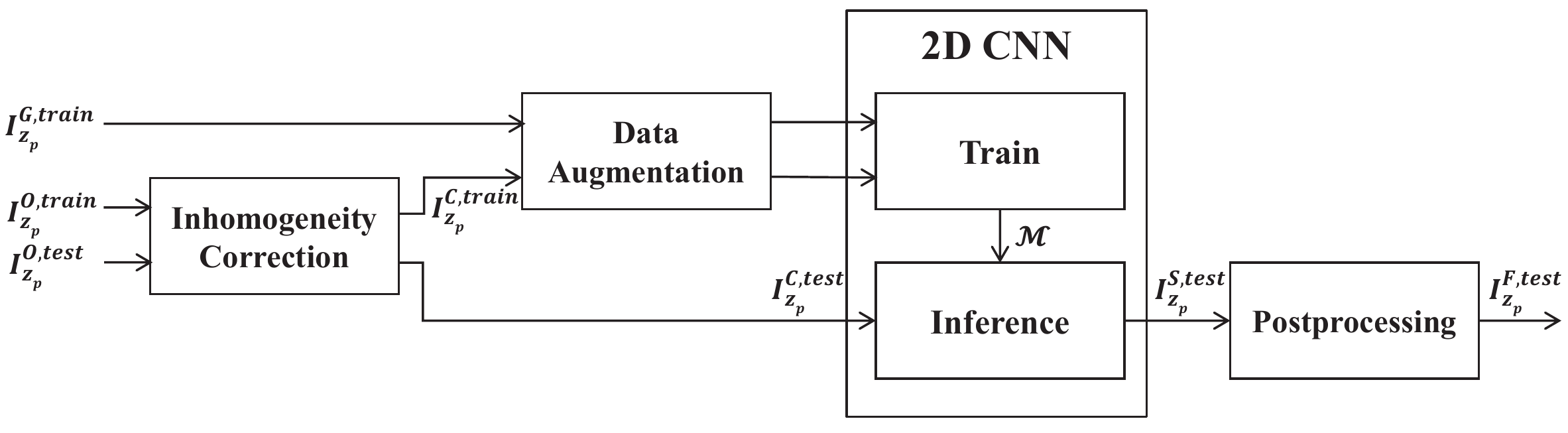}
	\caption{Block diagram of the proposed segmentation method for tubule segmentation} \label{fig:blockdiagram}
\end{figure}

Figure \ref{fig:blockdiagram} shows a block diagram of the proposed method. We denote a 3D image volume of size $X \times Y \times Z$ by $I$, and the $p^{th}$ focal plane image along the $z$-direction, of size $X \times Y$ pixels, by $I_{z_p}$ where $p \in \{1,\dots,Z\}$. We also denote the original training and test images in the $p^{th}$ focal plane by $I^{O,\,train}_{z_p}$ and $I^{O,\,test}_{z_p}$, respectively. In addition, $I^{G,\,train}_{z_p}$ and $I^{G,\,test}_{z_p}$ denote the groundtruth images that are used for training and testing that correspond to $I^{O,\,train}_{z_p}$ and $I^{O,\,test}_{z_p}$, respectively. Similarly, $I^{C,\,train}_{z_p}$ and $I^{C,\,test}_{z_p}$ denote inhomogeneity corrected training and test images, respectively. Lastly, $I^{S,\,test}_{z_p}$ denotes the binary segmentation mask generated by our proposed deep learning architecture and $I^{F,\,test}_{z_p}$ denotes the final segmentation outcome. For example, the $100^{th}$ original focal plane is denoted as $I^{O}_{z_{100}}$, its corresponding groundtruth image by $I^{G}_{z_{100}}$, the inhomogeneity corrected version by $I^{C}_{z_{100}}$, the binary segmentation mask  as $I^{S}_{z_{100}}$, and the final segmentation result by $I^{F}_{z_{100}}$, respectively.

As shown in Figure \ref{fig:blockdiagram}, our proposed network includes two stages: a training and an inference stage. During the training stage original training images ($I^{O,\,train}_{z_p}$) have their intensity inhomogeneities corrected ($I^{C,\,train}_{z_p}$) as a preprocessing step. Since fluorescence microscopy images suffer from intensity inhomogeneity due to non-uniform light attenuation, correcting intensity inhomogeneity helps improve final segmentation results. We then utilize both $I^{C,\,train}_{z_p}$ and $I^{G,\,train}_{z_p}$ as inputs to the data augmentation step to increase the number of training image pairs used for training the CNN model, $\mathcal{M}$. During the inference stage inhomogeneity correction is done on the test images ($I^{O,\,test}_{z_p}$) to obtain $I^{C,\,test}_{z_p}$. These $I^{C,\,test}_{z_p}$ are then used to segment tubules with the trained model $\mathcal{M}$.

\subsection{Intensity Inhomogeneity Correction}
Due to non-uniform intensities of fluorescence microscopy where center regions of the focal plane are generally brighter than boundary regions, simple intensity based segmentation methods failed to segment biological structures especially near image boundaries \cite{bib:SLee2017}. Our previous work \cite{bib:SLee2017} employed a multiplicative model where the original microscopy volume is modeled as
\begin{equation} \label{eq:model}
I^O = W\circ I^C + N.
\end{equation}
Here, $W$ and $N$ are a 3D weight array and a zero mean 3D Gaussian noise array, respectively, both of same size as the original microscopy volume. Specifically, $W$ represents weight values for each voxel location that accounts for the degree of intensity inhomogeneity. The $\circ$ operator denotes the Hadamard product representing voxelwise multiplication. 

The main idea of the multiplicative model is that an original volume, $I^O$, is modeled as the product of a 3D inhomogeneity field $W$ with a corrected volume $I^C$ and the product corrupted by additive 3D Gaussian noise $N$. In \cite{bib:SLee2017} an iterative technique to finding $W$ and then correcting for the intensity inhomogeneities based on this model is described.

Our proposed method uses this inhomogeneity correction technique as a preprocessing step for both training and inference. Examples of original and inhomogeneity corrected images are shown in Figure \ref{fig:orig} and Figure \ref{fig:inhomoCorrected}, respectively.

\begin{figure*}[htbp!]
	\centering
	\includegraphics[width=0.95\textwidth]{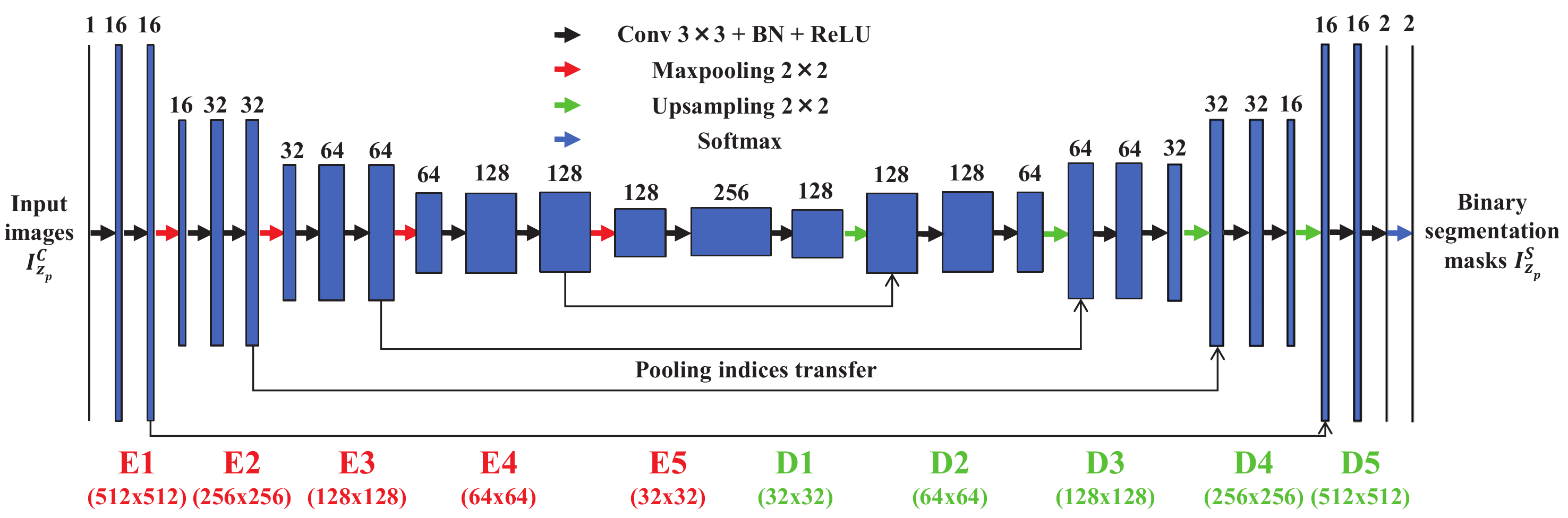}
	\caption{Proposed convolutional neural network architecture} \label{fig:CNN}
\end{figure*}

\subsection{Data Augmentation}
Our training data consists of paired images which are original microscopy images and corresponding manually annotated groundtruth images. Generating manually annotated groundtruth images is a time consuming process and thus impractical when generating large numbers of images. Data augmentation is typically used when the available training data size is relatively small to generate additional groundtruth images \cite{bib:UNet}. In this paper we utilize an elastic deformation to generate realistic tubular structures with different shapes and orientations. This allows the network to learn various deformed tubular structures, and is particularly useful for analysis of microscopy images especially for tubular structures that appear in varying shapes and orientations \cite{bib:UNet}. 

We used elastic deformation by employing a grid of control points located every $64$ pixels along the horizontal and vertical directions and displacing these control points randomly within $15$ pixels in each direction to generate a deformation field. The deformation field is used to deform the $p^{th}$ focal planes $I^{C,\,train}_{z_p}$ and $I^{G,\,train}_{z_p}$ by fitting 2D B-spline basis function to the grid followed by bicubic interpolation \cite{bib:Lorenz2012}. We generated $100$ random deformation fields for each image pairs and use them to generate $100$ deformed image pairs. Each deformed image is rotated $0^{\circ}$, $90^{\circ}$, $180^{\circ}$, $270^{\circ}$ to generate four sets of rotated images while preserving the original image size. Each rotated image is then flipped left and right to generate another two sets of images. In our experiment, we manually annotated five pairs of training data during the training stage. Since the elastic deformation uses $100$ deformations followed by four rotations and two flips for each deformed image, $4000$ pairs of images were generated for training.

\subsection{Convolutional Neural Network (CNN)}
The architecture of our convolutional neural network, shown in Figure \ref{fig:CNN}, consists of $5$ encoder layers denoted as $E1$ through $E5$ and $5$ decoder layers denoted as $D1$ through $D5$ that are serially connected followed by a softmax layer at the end. Each encoder layer consists of a $3 \times 3$ kernel with $1$ pixel padding to maintain same image size, a batch normalization step \cite{bib:BatchNorm} to perform image whitening, followed by a rectifier-linear unit (ReLU). 
The combination of convolution, batch normalization, and ReLU are performed twice at every encoder. Finally, maxpooling with a stride of $2$ is used to reduce dimensionality. This encoder scheme is similar to VGGNet \cite{bib:VGGNet} which shrinks the input dimensions but increases the number of filters in the deeper structures. In Figure \ref{fig:CNN}, each encoder's input dimension is indicated in red under the encoder layers. Also note that the number shown above each layer represents the number of filters utilized for training. For example, an input image of size  $512 \times 512 \times 1$ is resized to  $256 \times 256 \times 16$ at the input to the E2 layer. As the image passes through the all encoder layers, its X and Y dimensions shrink to $32$, respectively, but number of filters utilized increases to $256$. Therefore, the input to the first decoder layer is of dimension $32 \times 32 \times 256$. 

Conversely, each decoder is comprised of two $3 \times 3$ kernels with $1$ pixel padding, batch normalization, and ReLU. Instead of a maxpooling layer, the decoder has an unmaxpooling layer to upsample the data to increase dimensionality. Note that this upsampling process is a reconstruction process. To achieve better upsampling maxpooling indices from each encoder layer are recorded and transferred to the corresponding same size unmaxpooling layer ($E1 \rightarrow D5, \dots, E4 \rightarrow D2$). At the end of the encoder-decoder structure, a softmax classifier layer is utilized to determine whether  each pixel location belongs to a tubule or background using a probability map. Note that the output of the softmax layer is of size $512 \times 512 \times 2$ because the final output includes two probability maps corresponding to the two classes: tubule or background. These probability maps are thresholded at $0.5$ to produce binary segmentation masks.

During the training stage augmented training images ($I^{C,\,train}_{z_p}$) are randomly selected and used to train the model $\mathcal{M}$ for each iteration. The segmentation mask is compared with the corresponding groundtruth ($I^{G,\,train}_{z_p}$) and a loss value is obtained for each iteration. 
Here, we use a 2D cross entropy loss function that is minimized using stochastic gradient descent (SGD) with a fixed learning rate and a momentum. During the inference stage we use the trained model $\mathcal{M}$ with test images ($I^{C,\,test}_{z_p}$) to obtain binary segmentation masks ($I^{S,\,test}_{z_p}$). During the postprocessing step we clean up objects less than $\gamma$ pixels from $I^{S,\,test}_{z_p}$ followed by a hole filling operation to obtain final segmentation results ($I^{F,\,test}_{z_p}$). Note that the hole filling operation assigns a background pixel to a tubule pixel if the background pixel's $4$ neighborhood pixels are all tubule pixels.

\section{Experimental Results}
\label{sec:exp}

\begin{figure}[ht!]
\vspace{-0.1in}
	\centering
	\subfloat[$I^O_{z_{100}}$]
		 {\label{fig:orig}\includegraphics[width=0.14\textwidth]{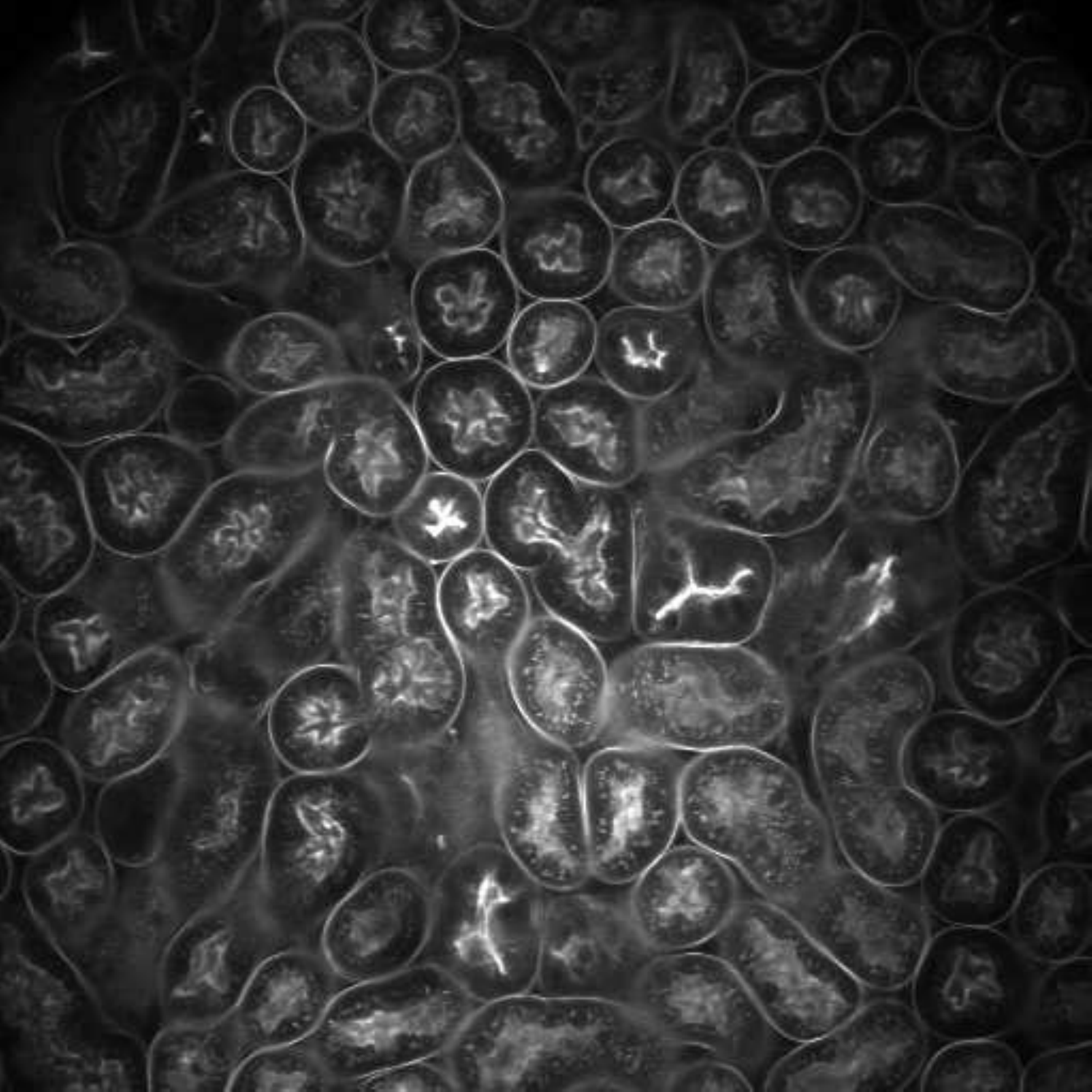}}
	\,\,\,
	\subfloat[$I^C_{z_{100}}$]
		 {\label{fig:inhomoCorrected}\includegraphics[width=0.14\textwidth]{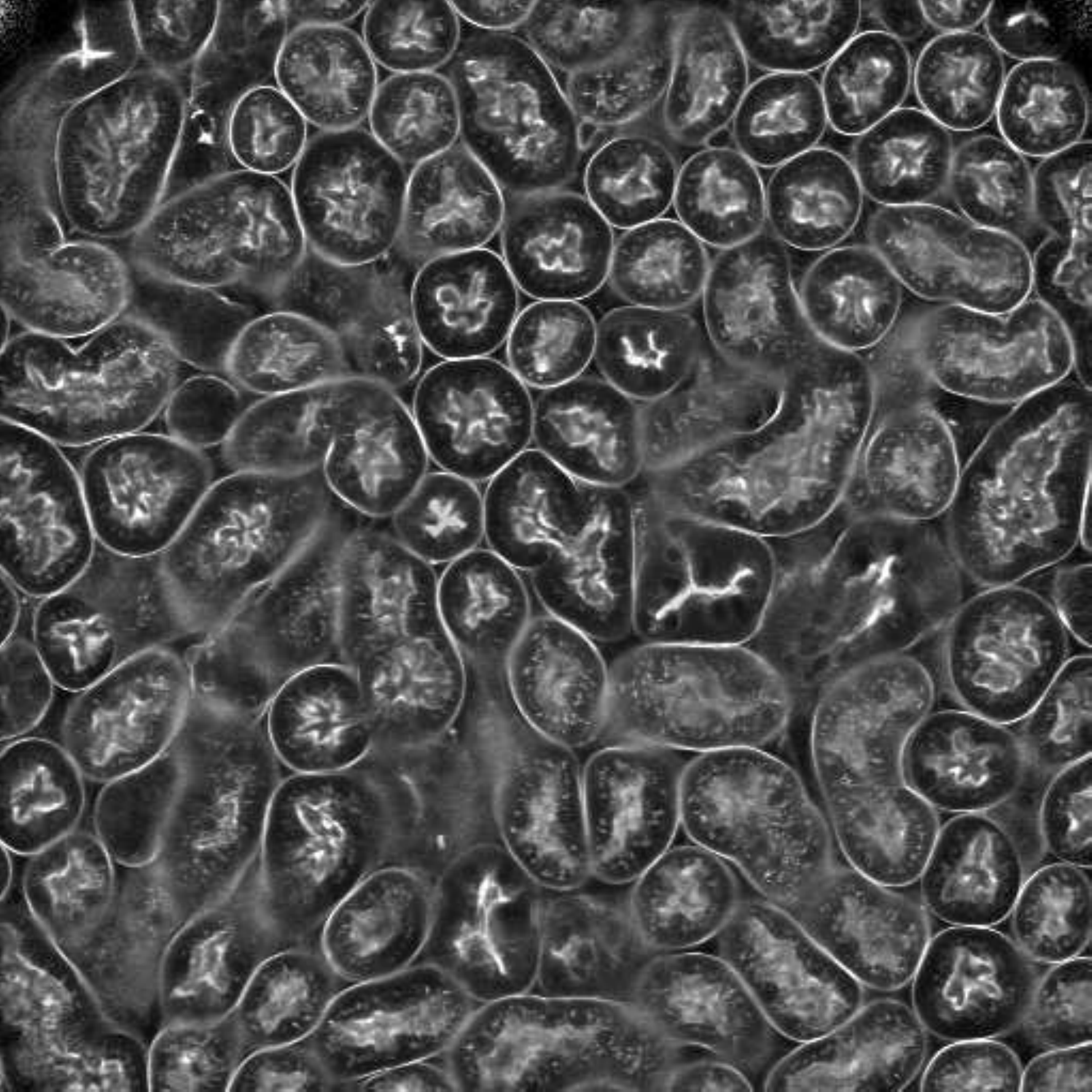}}
	\,\,\,
 	\subfloat[$I^G_{z_{100}}$]
		 {\label{fig:groundtruth}\includegraphics[width=0.14\textwidth]{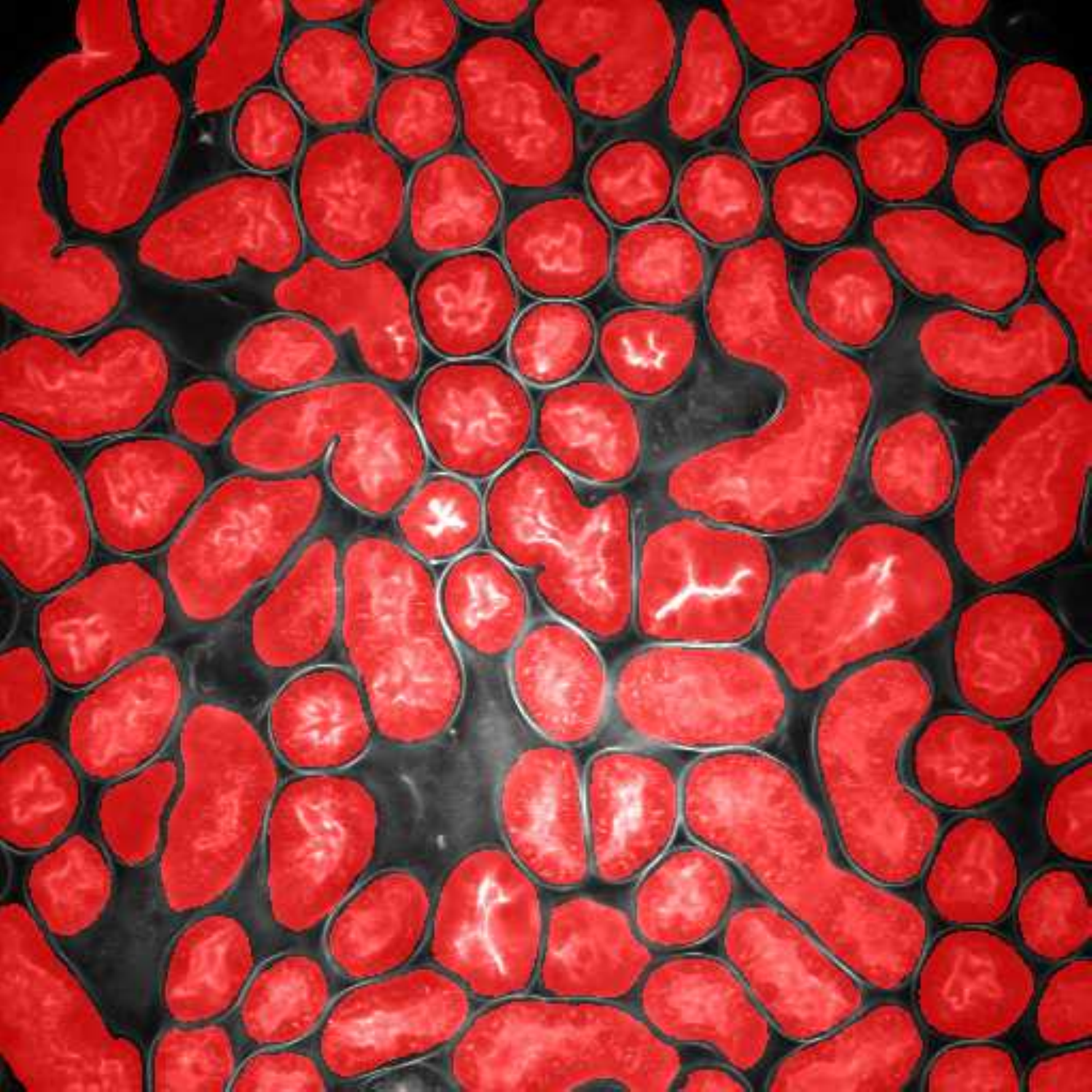}}\\
\vspace{-0.2in}
	\,\,\,
	\subfloat[\textit{3Dac}]
		 {\label{fig:3Dac_segOverlaidOrig}\includegraphics[width=0.14\textwidth]{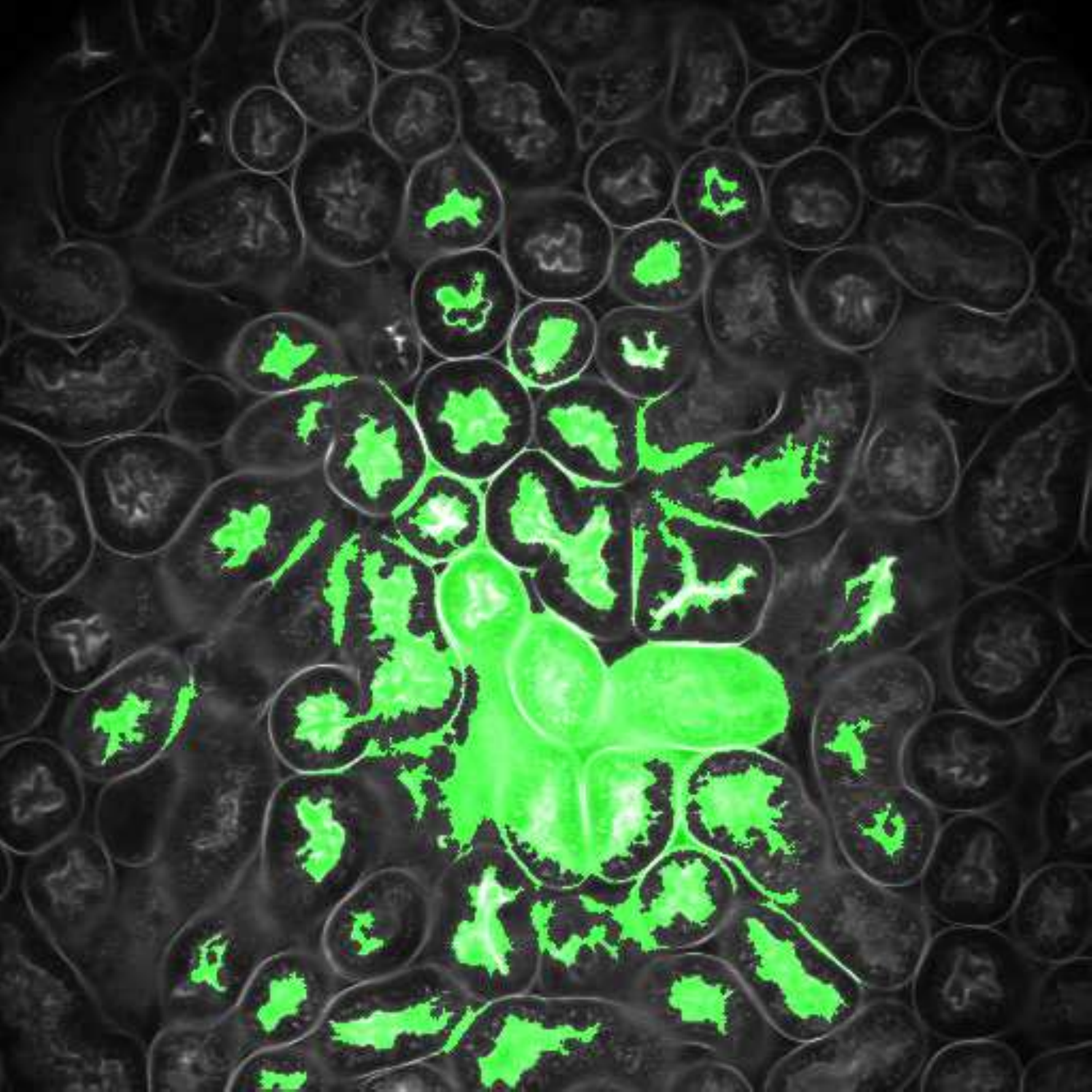}}
	\,\,\,
	\subfloat[\textit{3DacIC}]
		 {\label{fig:3DacIC_segOverlaidOrig}\includegraphics[width=0.14\textwidth]{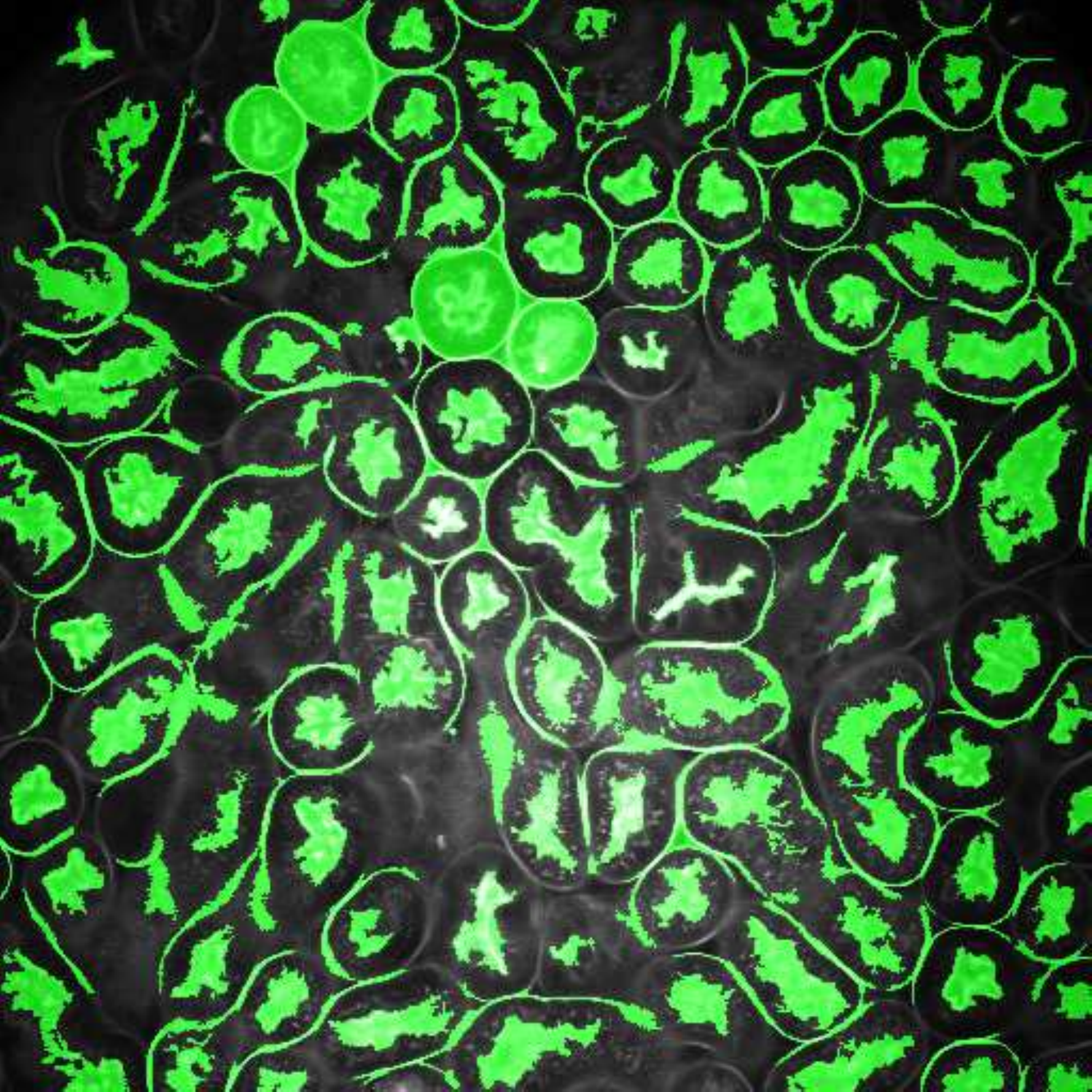}} 	
	\,\,\,
	\subfloat[\textit{3Dsquassh}]
		 {\label{fig:3Dsquassh_segOverlaidOrig}\includegraphics[width=0.14\textwidth]{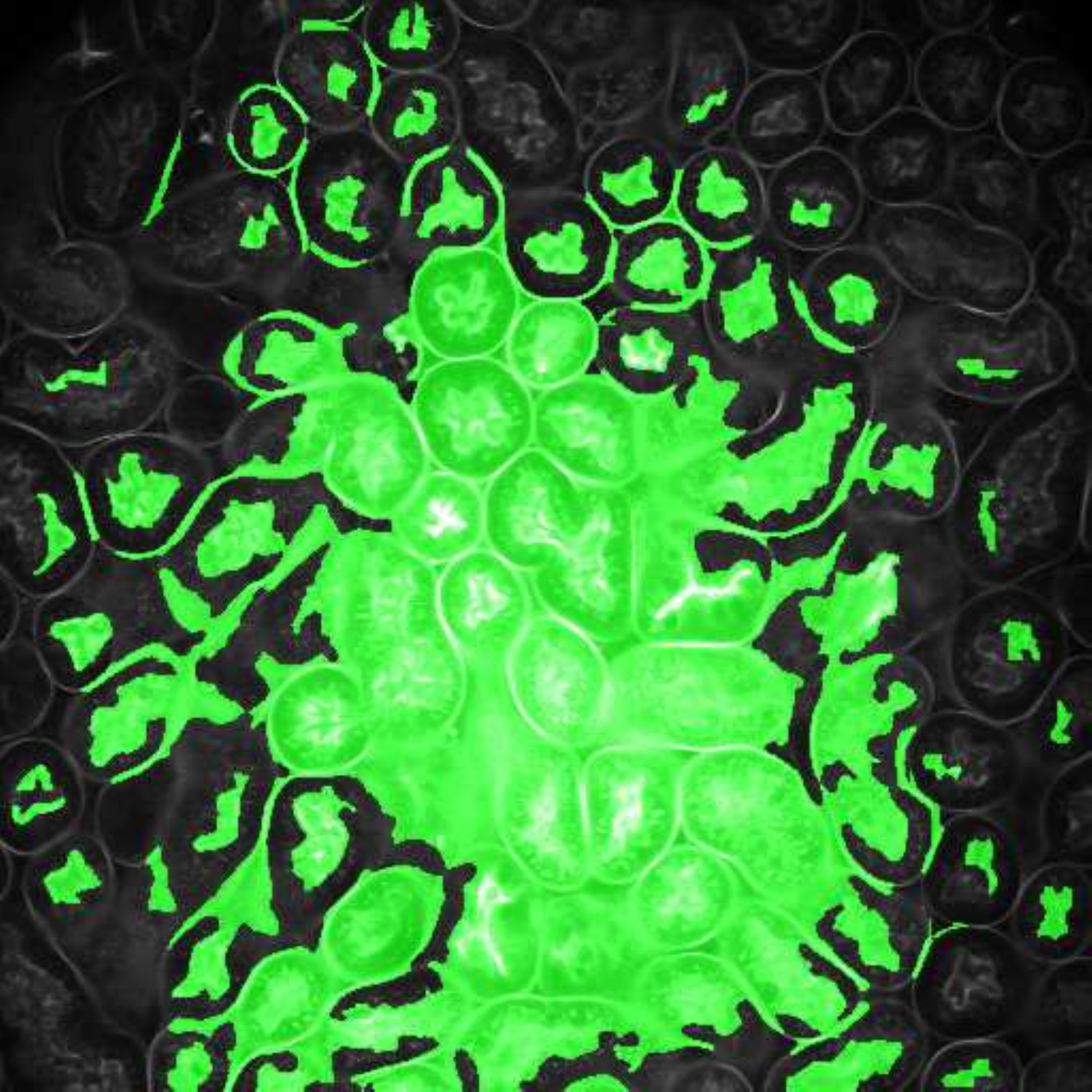}} \\
\vspace{-0.2in}
	\,\,\,
	\subfloat[\textit{Ellipse Fitting}]	
		 {\label{fig:ellipsefitting_segOverlaidOrig}\includegraphics[width=0.14\textwidth]{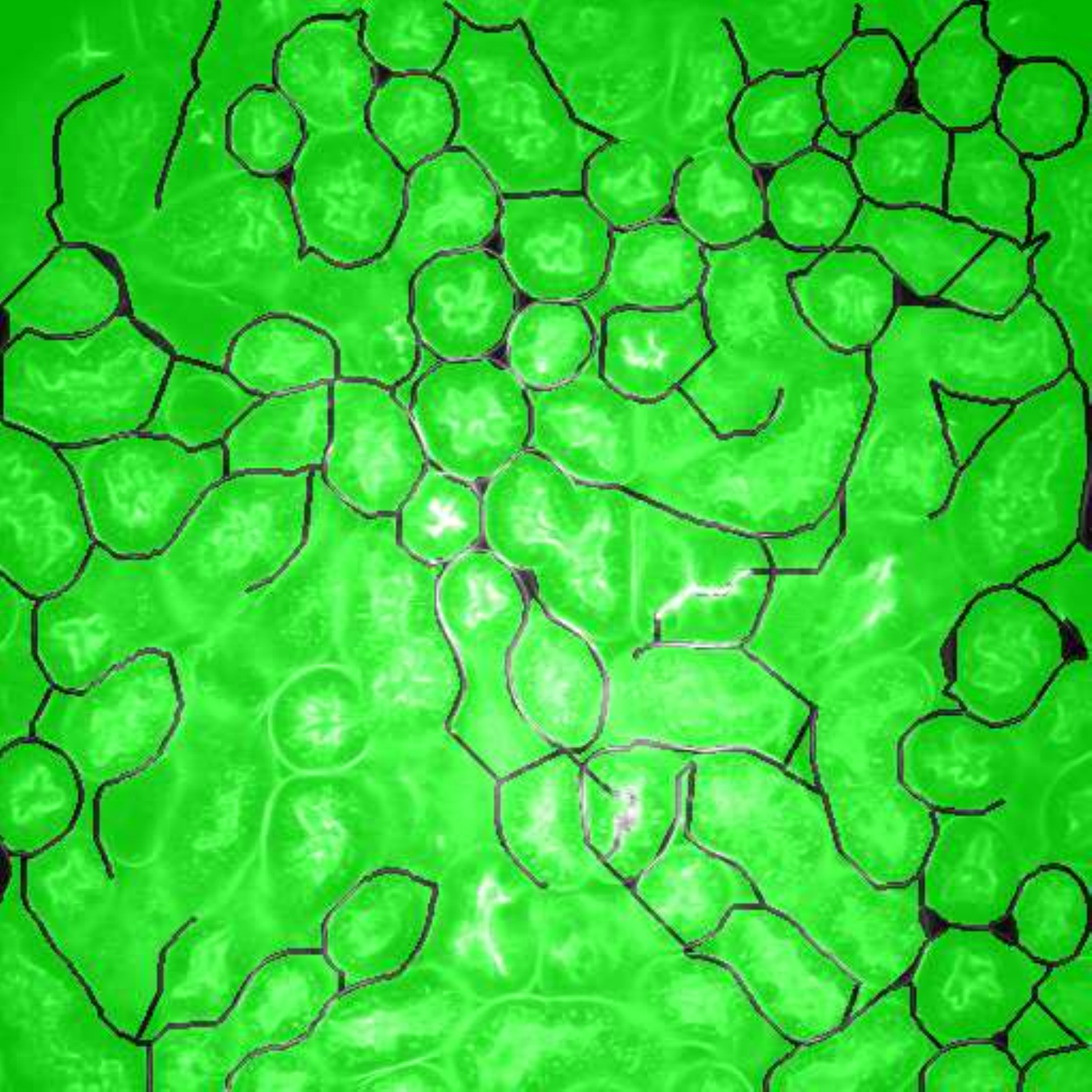}}
		 \,\,\,
	\subfloat[\textit{Jelly Filling}]
		 {\label{fig:jellyfilling_segOverlaidOrig}\includegraphics[width=0.14\textwidth]{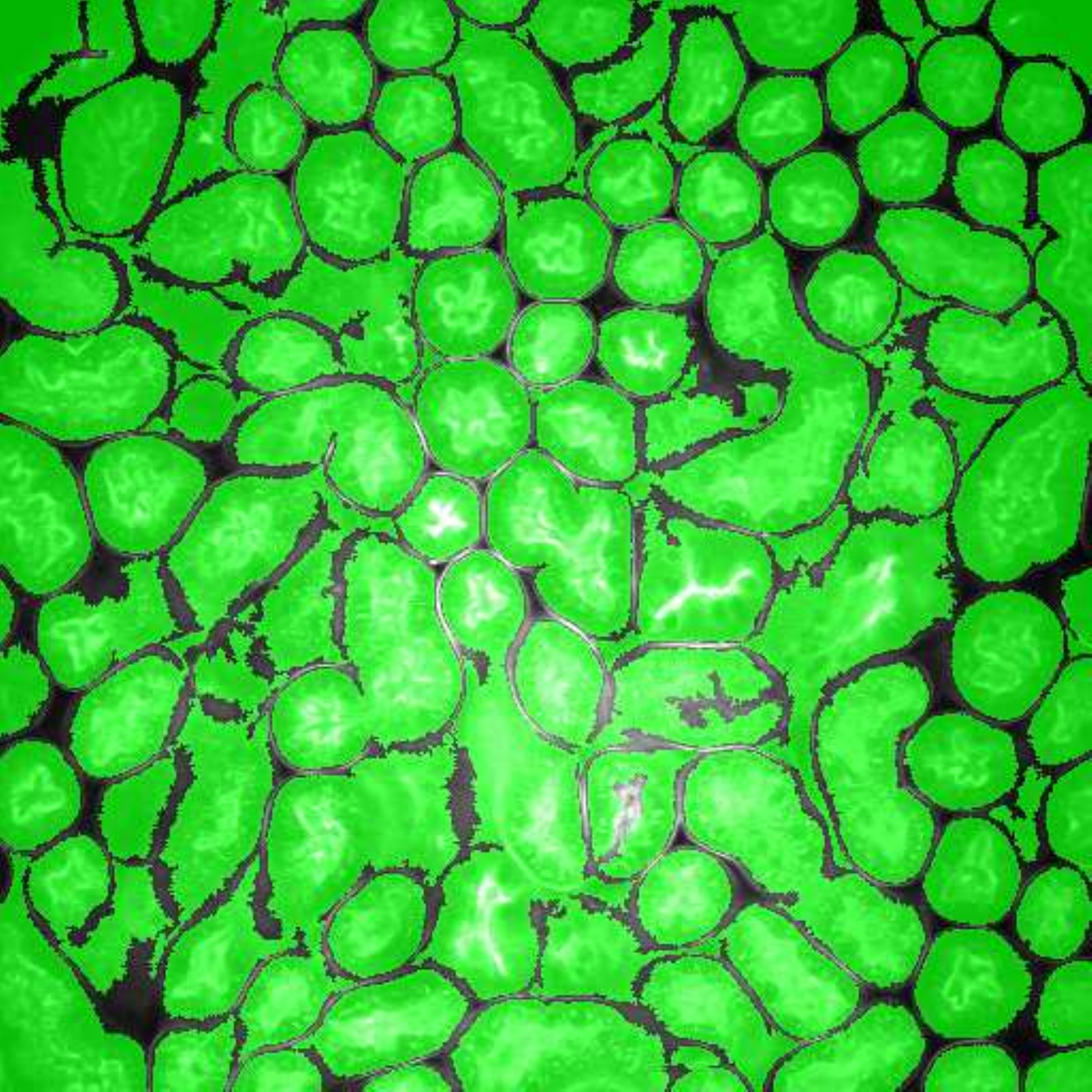}}
		 \,\,\,
	\subfloat[\textit{Steerable Filter}]
		 {\label{fig:steerablefilter_segOverlaidOrig}\includegraphics[width=0.14\textwidth]{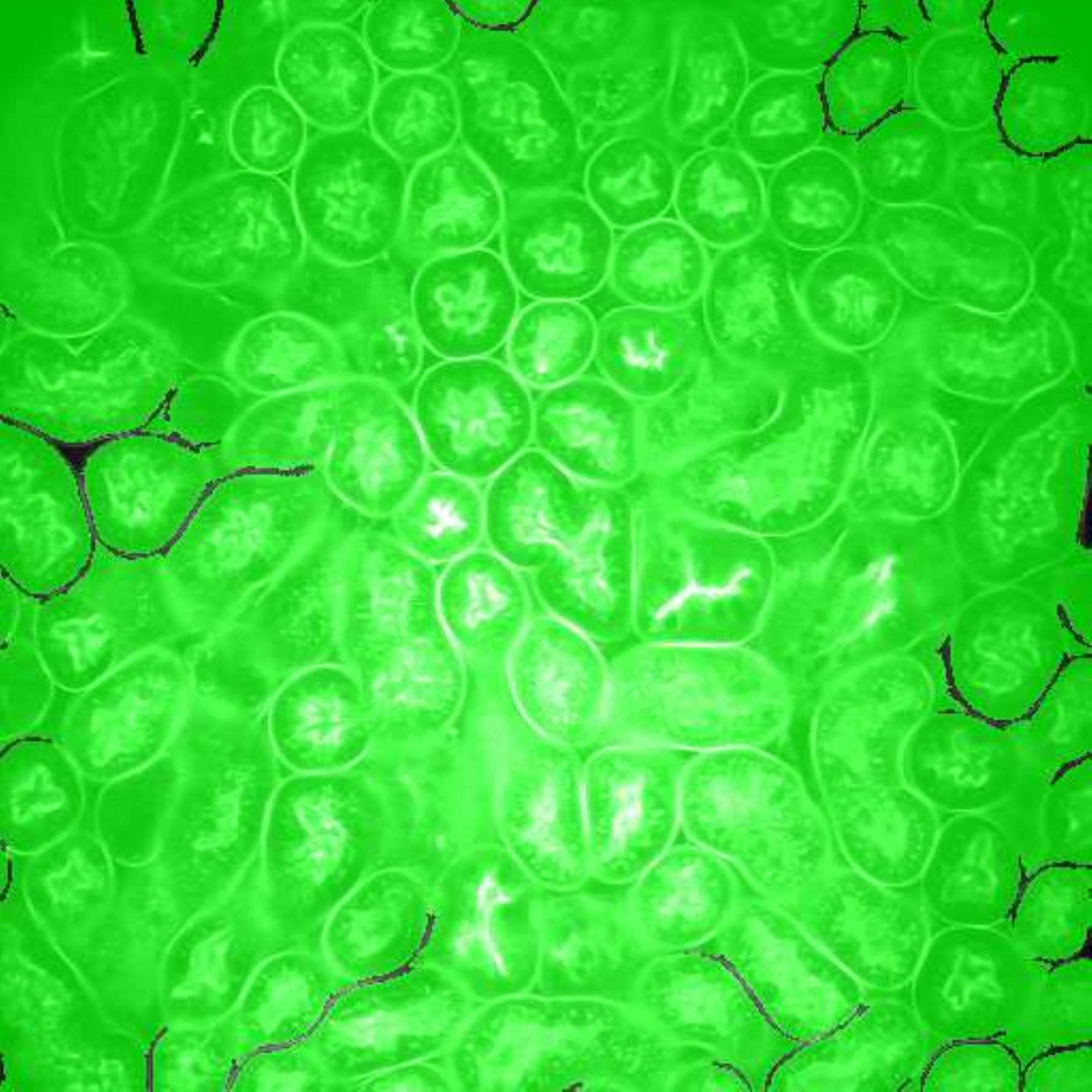}}\\
	 	\quad
	\subfloat[\textit{2DCNN}]
		 {\label{fig:2DCNN_segOverlaidOrig}\includegraphics[width=0.14\textwidth]{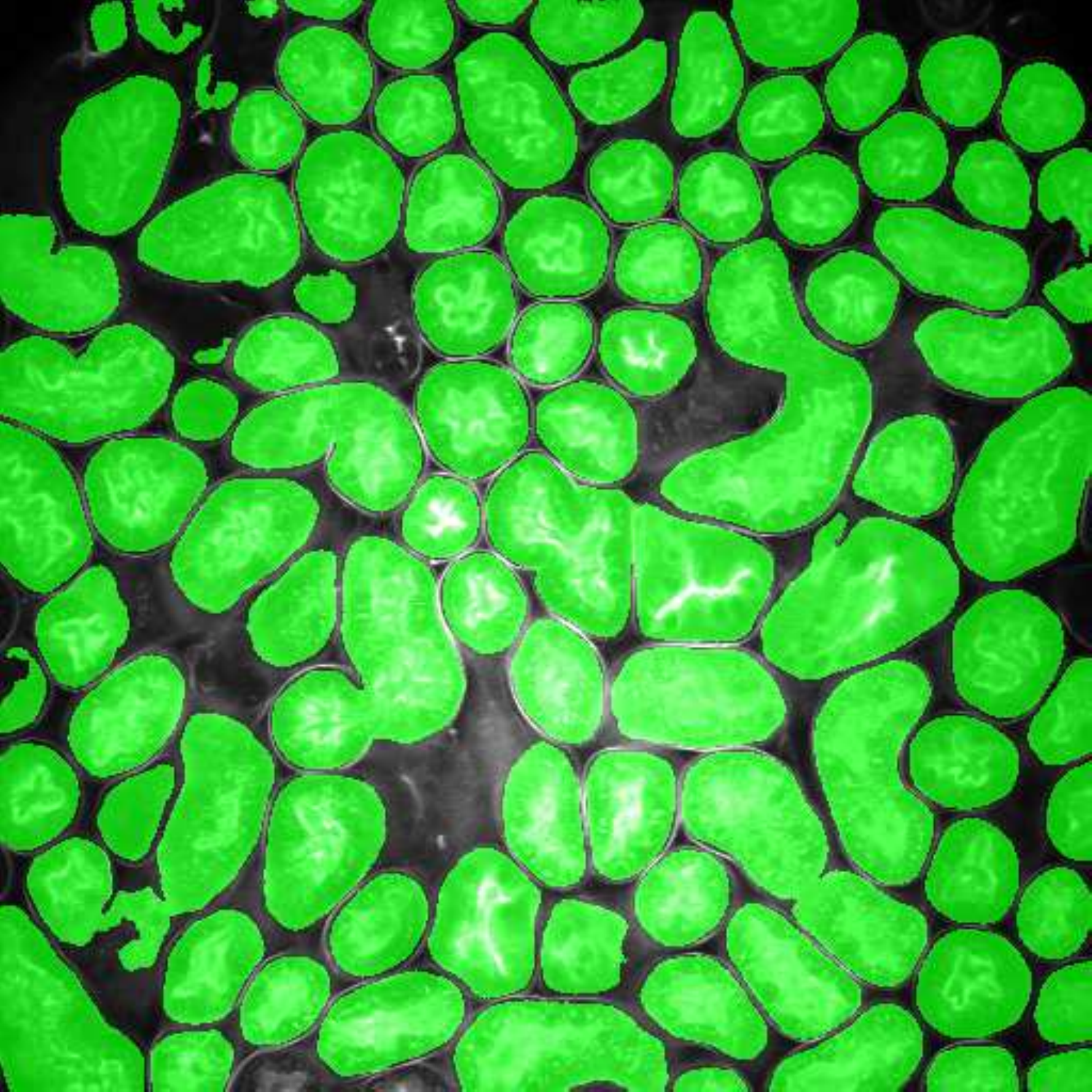}}
		\quad
		\captionsetup[subfloat]{indent=0.02\textwidth, width=0.10\textwidth}
		\subfloat[\textit{2DCNNIC} (Proposed)]
		 {\label{fig:2DCNNIC_segOverlaidOrig}\includegraphics[width=0.14\textwidth]{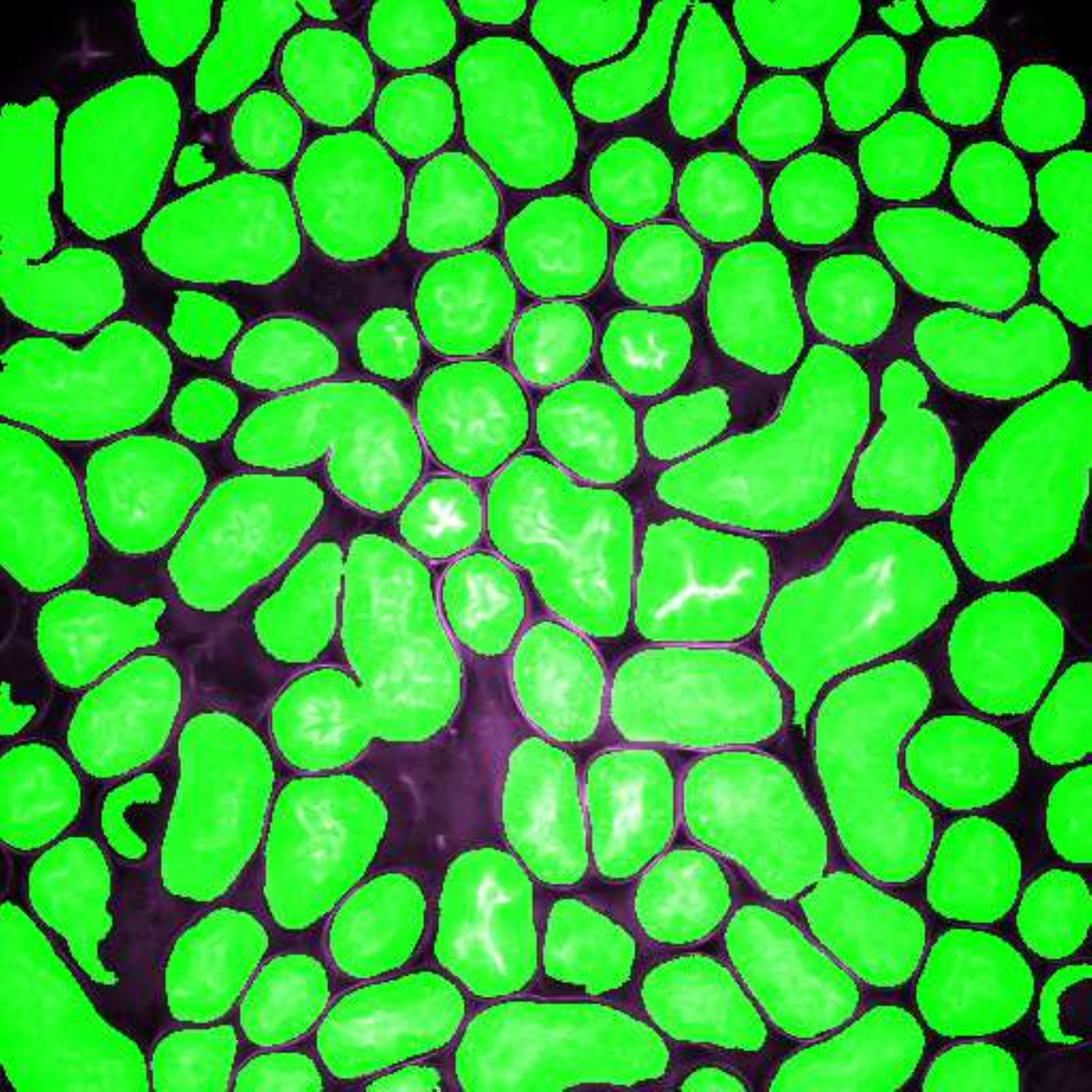}}\\
\caption{Segmentation results obtained by the proposed method and other methods as well as the corresponding groundtruth data for the $100^{th}$ image ($I_{z_{100}}$) in $Dataset-I$. Segmentation results are highlighted in green and corresponding groundtruth in red.}
\label{fig:visComparison1} 
\end{figure}

The performance of our proposed method was tested on two different datasets:\footnote{$Dataset-I$ and $II$ were provided by Malgorzata Kamocka of the Indiana Center for Biological Microscopy.}
$Dataset-I$ and $II$. $Dataset-I$ is comprised of $Z = 512$ grayscale images, each of size $X \times Y = 512 \times 512$ pixels, whereas $Dataset-II$ consists of $Z = 821$ grayscale images, each of size $X \times Y = 640 \times 640$ pixels. We selected five different images from $Dataset-I$ and generated corresponding manually annotated groundtruth images  to train model $\mathcal{M}$. Our deep learning architecture was implemented in Torch $7$ \cite{bib:Torch} using a fixed learning rate $10^{-5}$ and a momentum of $0.9$. As indicated, $4000$ pairs of images were generated by the elastic deformation, rotations, and flips using these five pairs of images. Note that each training data was used as a batch so that $4000$ iterations were performed per epoch. We used $200$ epochs for training our proposed network. In addition, $\gamma = 100$ was used for the removal of small objects. The performance of the proposed method was evaluated using manually annotated groundtruth images ($I^{G,\,test}_{z_p}$) at different depths in $Dataset-I$ that were never used during the training stage. For visual evaluation and comparison segmentation results of $I_{z_{100}}$ in $Dataset-I$ using various techniques are presented in Figure \ref{fig:visComparison1}.

\subsection{Qualitative Evaluation}
The first row in Figure \ref{fig:visComparison1} displays an original microscopy image ($I^O_{z_{100}}$), its inhomogeneity corrected version ($I^C_{z_{100}}$), and manually delineated groundtruth ($I^G_{z_{100}}$), respectively. For brevity we have omitted the superscript $test$ in the notation. The second row shows segmentation results of various 3D methods such as 3D region-based active contours \cite{bib:Lorenz2013} (\textit{3Dac}), 3D active contours with inhomogeneity correction \cite{bib:SLee2017} (\textit{3DacIC}), and 3D Squassh  presented in \cite{bib:Paul2013} (\textit{3Dsquassh}). Similarly, the third row portrays various segmentation methods particularly designed for tubular structure segmentation such as ellipse fitting method presented in \cite{bib:SLee2015} (\textit{Ellipse Fitting}), the Jelly filling method in \cite{bib:Gadgil2016b} (\textit{Jelly Filling}), and tubule segmentation using steerable filter \cite{bib:DHo2017a} (\textit{Steerable Filter}). Finally, the last row shows segmentation results of our proposed CNN architecture without inhomogeneity correction \cite{bib:CFu2017} (\textit{2DCNN}) and with inhomogeneity correction (\textit{2DCNNIC}).

For visual comparison we highlighted groundtruth regions in red, segmented tubule regions in green, and background in black. As observed in Figure \ref{fig:visComparison1}, our proposed method appeared to perform better than the other six methods shown in the second and third rows by distinguishing tubules and was similar performance to \textit{2DCNN}. Note that since some methods such as \textit{Ellipse Fitting}, \textit{Jelly Filling}, and \textit{Steerable Filter} only segmented boundaries of tubule structures, tubule interiors were filled in order to perform a fair comparison using connected components with a $4$-neighborhood systems. Also, based on the assumption that tubule regions should contain lumen, if a filled region contained lumen pixel, the region was identified as a tubule region. However, if a filled region did not contain any lumen pixels, the region was considered as a background region.

\begin{table*}[ht!]
\centering
\renewcommand{\tabcolsep}{2pt}
{
\begin{tabular}{|c||c|c|c|c|c|c||c|c|c|c|c|c|}
\hline
& \multicolumn{6}{c||}{$I_{z_{100}}$ of the $Dataset-I$} & \multicolumn{6}{c|}{$I_{z_{200}}$ of the $Dataset-I$} \\
\hline
{Method} & {PA} & {Type-I} & {Type-II} & {F1} & {OD} & {OH} & {PA} & {Type-I} & {Type-II} & {F1} & {OD} & {OH} \\
\hline
\textit{3Dac} \cite{bib:Lorenz2013} & 37.74\% & \textbf{3.31\%} & 58.95\% & 0.90\% & 20.86\% & 95.72
 & 38.98\% & \textbf{2.72\%} & 58.30\% & 1.83\% & 20.74\% & 125.69 \\
\hline
\textit{3DacIC} \cite{bib:SLee2017} & 42.92\% & 8.06\% & 49.02\% & 0.86\% & 36.45\% & 35.07
 & 44.58\% & 4.84\% & 50.59\% & 0.00\% & 38.02\% & 30.41 \\
\hline
\textit{3Dsquassh} \cite{bib:Paul2013} & 47.02\% & 11.80\% & 41.18\% & 1.83\% & 11.64\% & 223.34
 & 48.37\% & 9.55\% & 42.09\% & 1.94\% & 14.34\% & 181.58  \\
\hline
\textit{Ellipse Fitting} \cite{bib:SLee2015} & 76.17\% & 22.79\% & 1.04\% & 61.15\% & 47.10\% & 144.28
 & 76.11\% & 22.98\% & 0.91\% & 48.48\% & 29.34\% & 303.34 \\
\hline
\textit{Jelly Filling} \cite{bib:Gadgil2016b} & 83.91\% & 13.36\% & 2.73\% & 81.82\% & 71.58\% & 52.93
 & 81.76\% & 15.38\% & 2.86\% & 74.53\% & 60.73\% & 76.26 \\
\hline
\textit{Steerable Filter} \cite{bib:DHo2017a} & 70.98\% & 28.98\% & \textbf{0.04\%} & 9.90\% & 5.32\% & 455.83
 & 71.00\% & 28.97\% & \textbf{0.03\%} & 4.12\% & 4.00\% & 521.83 \\
\hline
\textit{2DCNN} \cite{bib:CFu2017} & \textbf{90.57\%} & 5.25\% & 4.17\% & 91.49\% & 90.09\% & 13.28
 & 86.92\% & 5.64\% & 7.44\% & 86.96\% & 87.10\% & 16.80 \\
\hline
\textit{2DCNNIC} &  \multirow{2}{*}{90.04\%} &  \multirow{2}{*}{6.44\%} &  \multirow{2}{*}{3.52\%} &  \multirow{2}{*}{\textbf{92.63\%}} &  \multirow{2}{*}{\textbf{90.12\%}} &  \multirow{2}{*}{\textbf{11.95}}
 &  \multirow{2}{*}{\textbf{88.66\%}} & \multirow{2}{*}{5.77\%} &  \multirow{2}{*}{5.57\%} & \multirow{2}{*}{\textbf{90.61\%}} & \multirow{2}{*}{\textbf{89.65\%}} &  \multirow{2}{*}{\textbf{11.76}} \\
(Proposed) &   &   &   &   &   &   &   &   &   &   &   &  \\
\hline
\end{tabular}
}
\vspace{0.05in}
\caption{Quantitative evaluation of the proposed method and other known methods in terms of Pixel Accuracy (PA), Type-I error, Type-II error, F1 score, Dice Index (OD), and Hausdorff Distance (OH)}
\label{tab:comp1}
\end{table*}

\begin{figure}[ht!]
\vspace{-0.1in}
	\centering
	\subfloat[\textit{3Dac}]
		 {\label{fig:3Dac_segOverlaidGT}\includegraphics[width=0.14\textwidth]{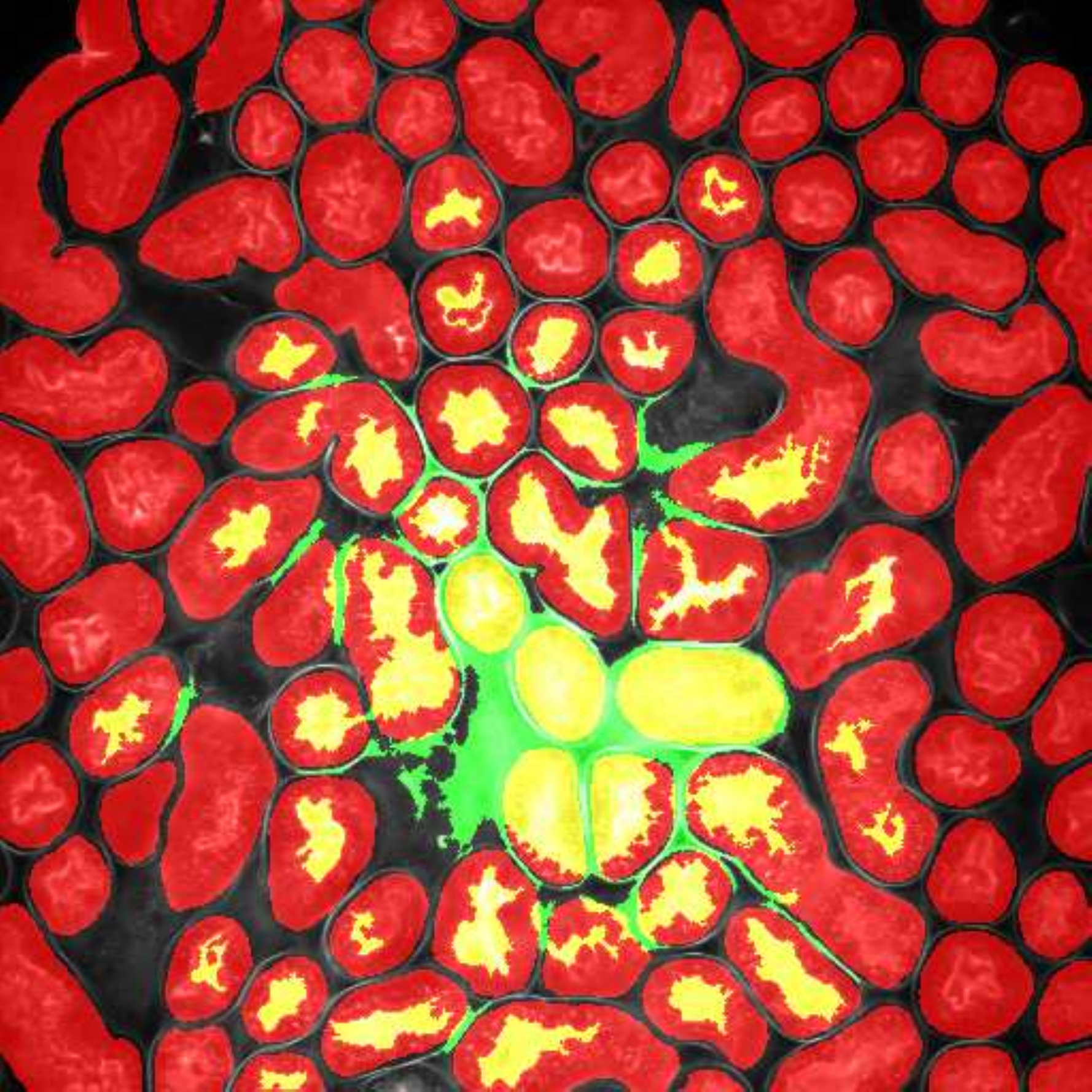}}
	\,\,\,
	\subfloat[\textit{3DacIC}]
		 {\label{fig:3DacIC_segOverlaidGT}\includegraphics[width=0.14\textwidth]{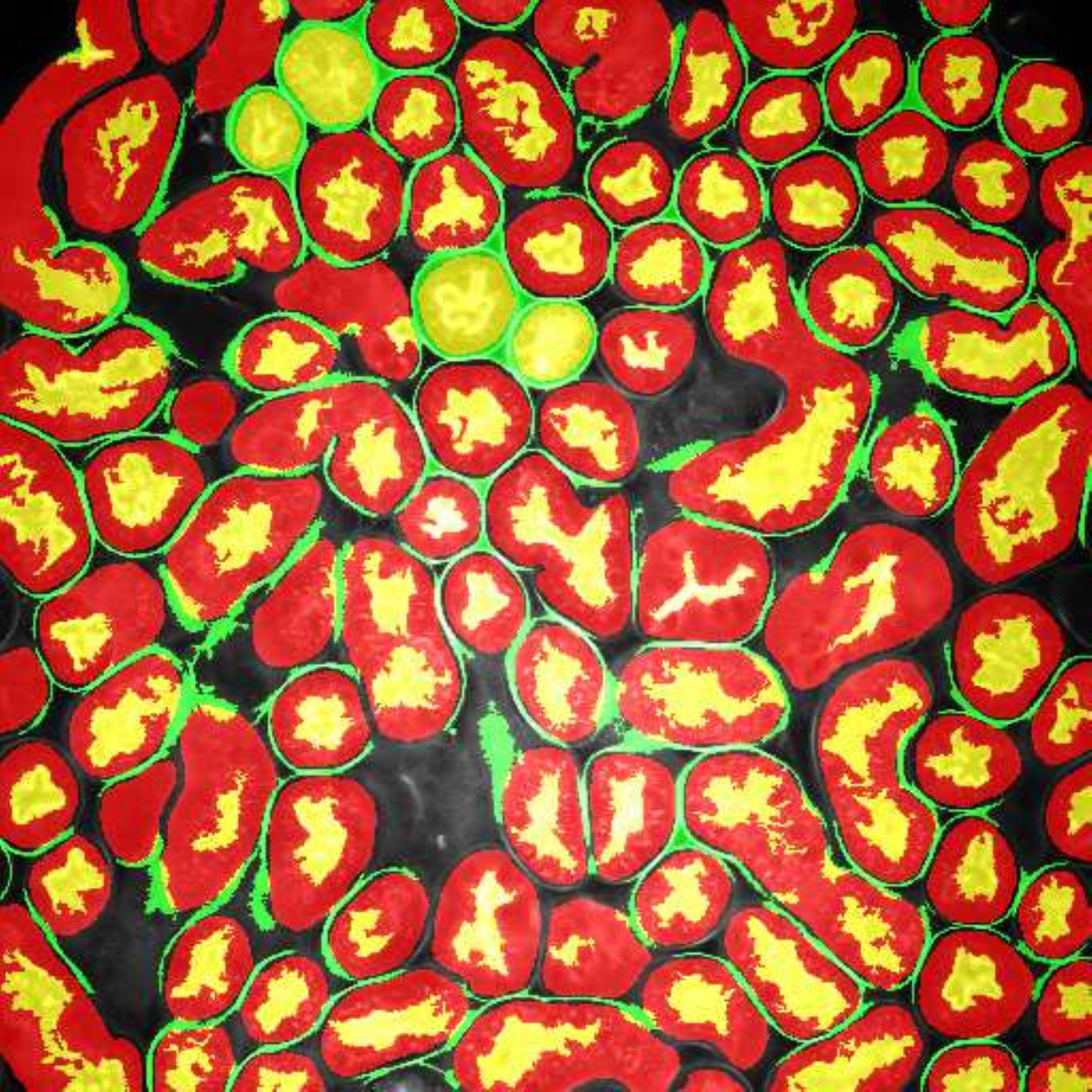}}
	\,\,\,
 	\subfloat[\textit{3Dsquassh}]
		 {\label{fig:3Dsquassh_segOverlaidGT}\includegraphics[width=0.14\textwidth]{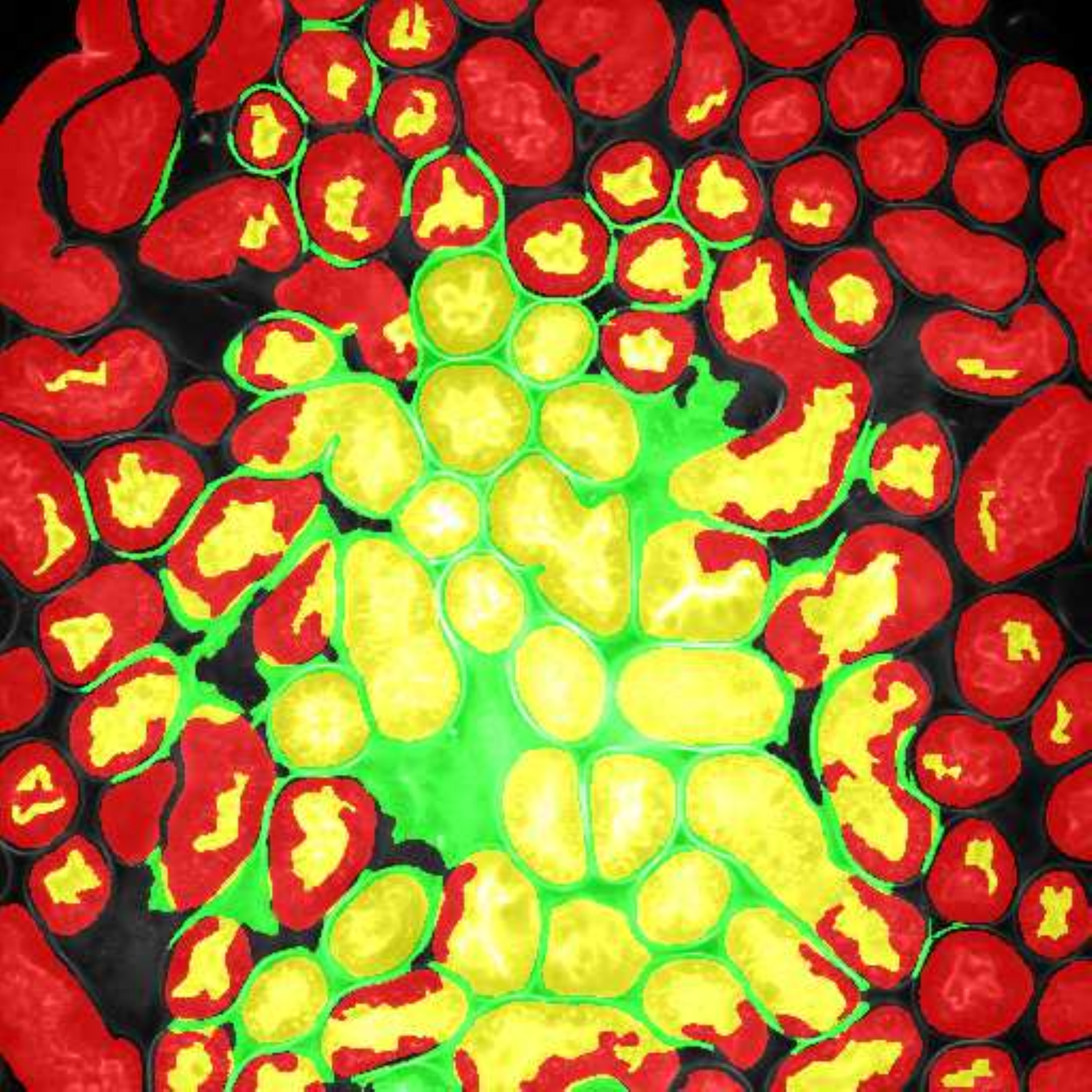}}\\
\vspace{-0.2in}
	\,\,\,
	\subfloat[\textit{Ellipse Fitting}]
		 {\label{fig:ellipsefitting_segOverlaidGT}\includegraphics[width=0.14\textwidth]{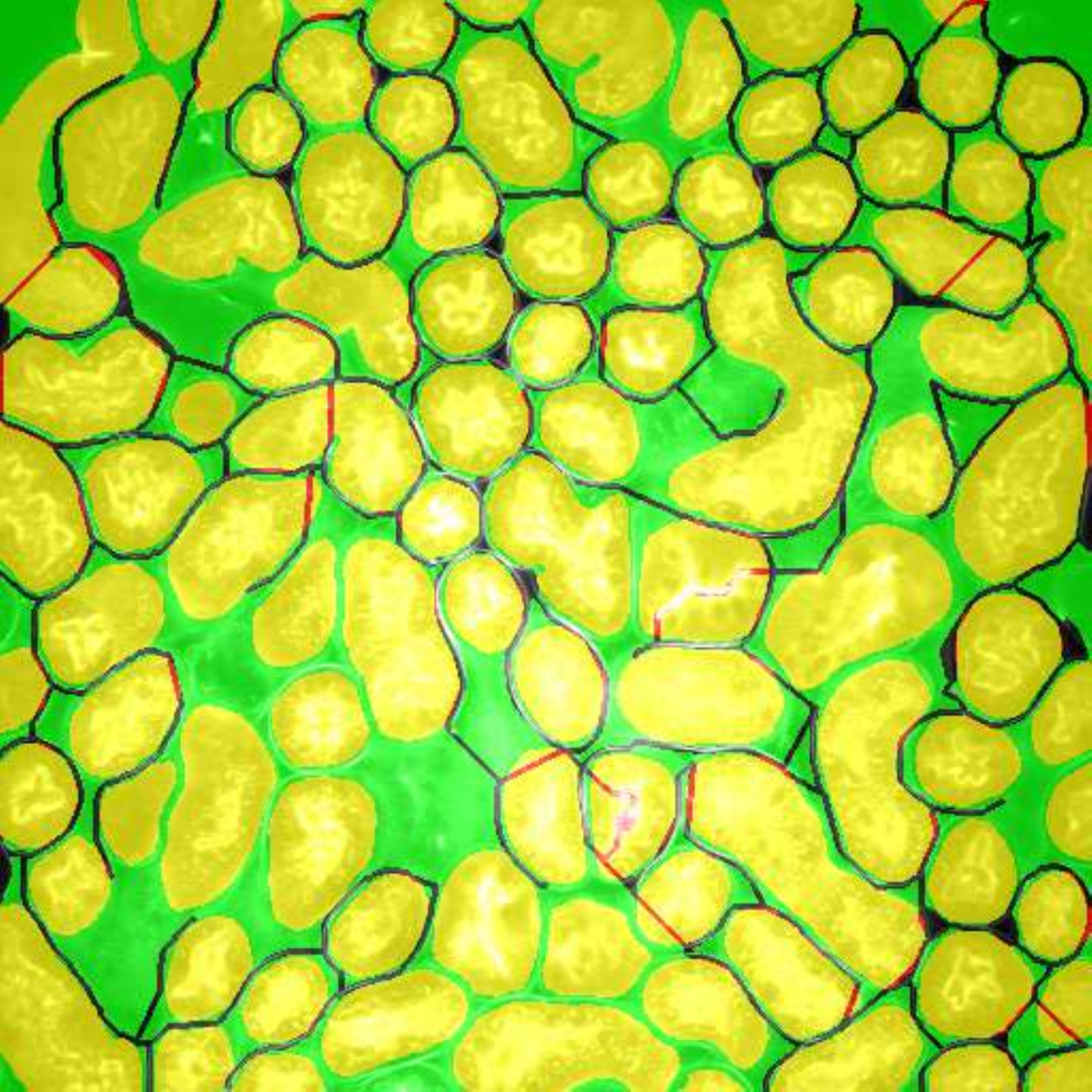}}
	\,\,\,
	\subfloat[\textit{Jelly Filling}]
		 {\label{fig:jellyfilling_segOverlaidGT}\includegraphics[width=0.14\textwidth]{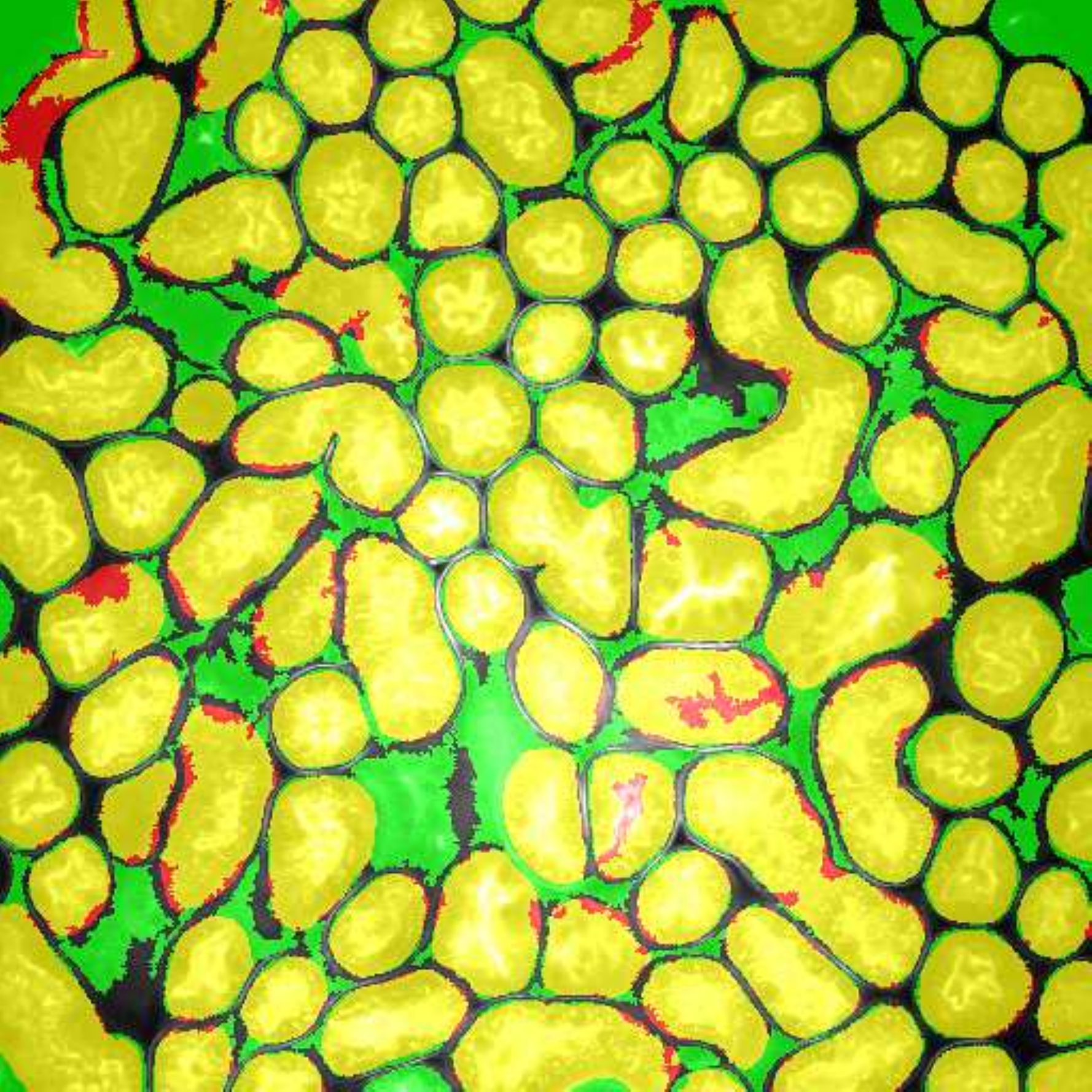}}	
	\,\,\,
	\subfloat[\textit{Steerable Filter}]
		 {\label{fig:steerablefilter_segOverlaidGT}\includegraphics[width=0.14\textwidth]{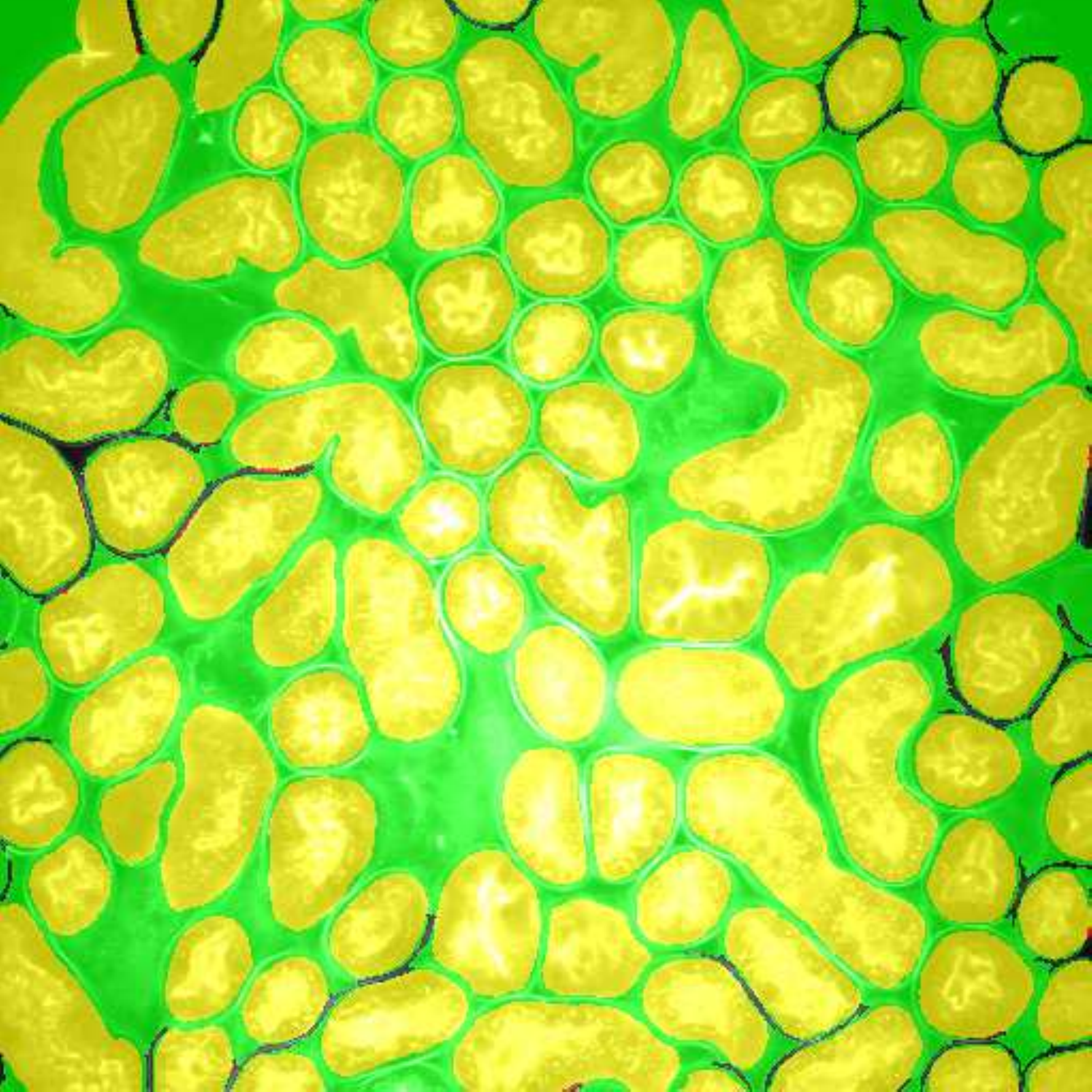}}\\
\vspace{-0.1in}
	\subfloat[\textit{2DCNN}]
		 {\label{fig:2DCNN_segOverlaidGT}\includegraphics[width=0.14\textwidth]{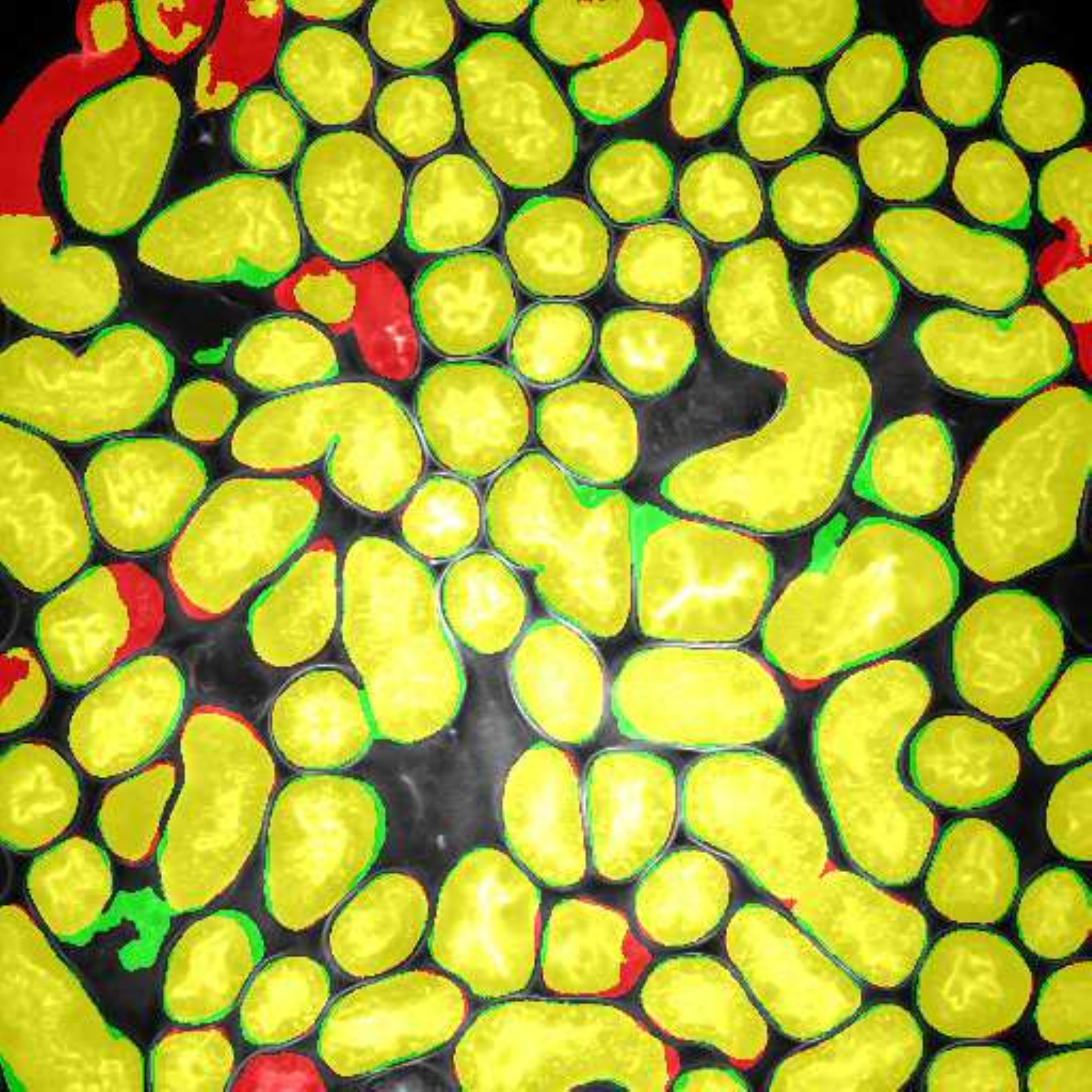}}
	\qquad
	\captionsetup[subfloat]{indent=0.02\textwidth, width=0.10\textwidth}
	\subfloat[\textit{2DCNNIC} (Proposed)]
		 {\label{fig:2DCNNIC_segOverlaidGT}\includegraphics[width=0.14\textwidth]{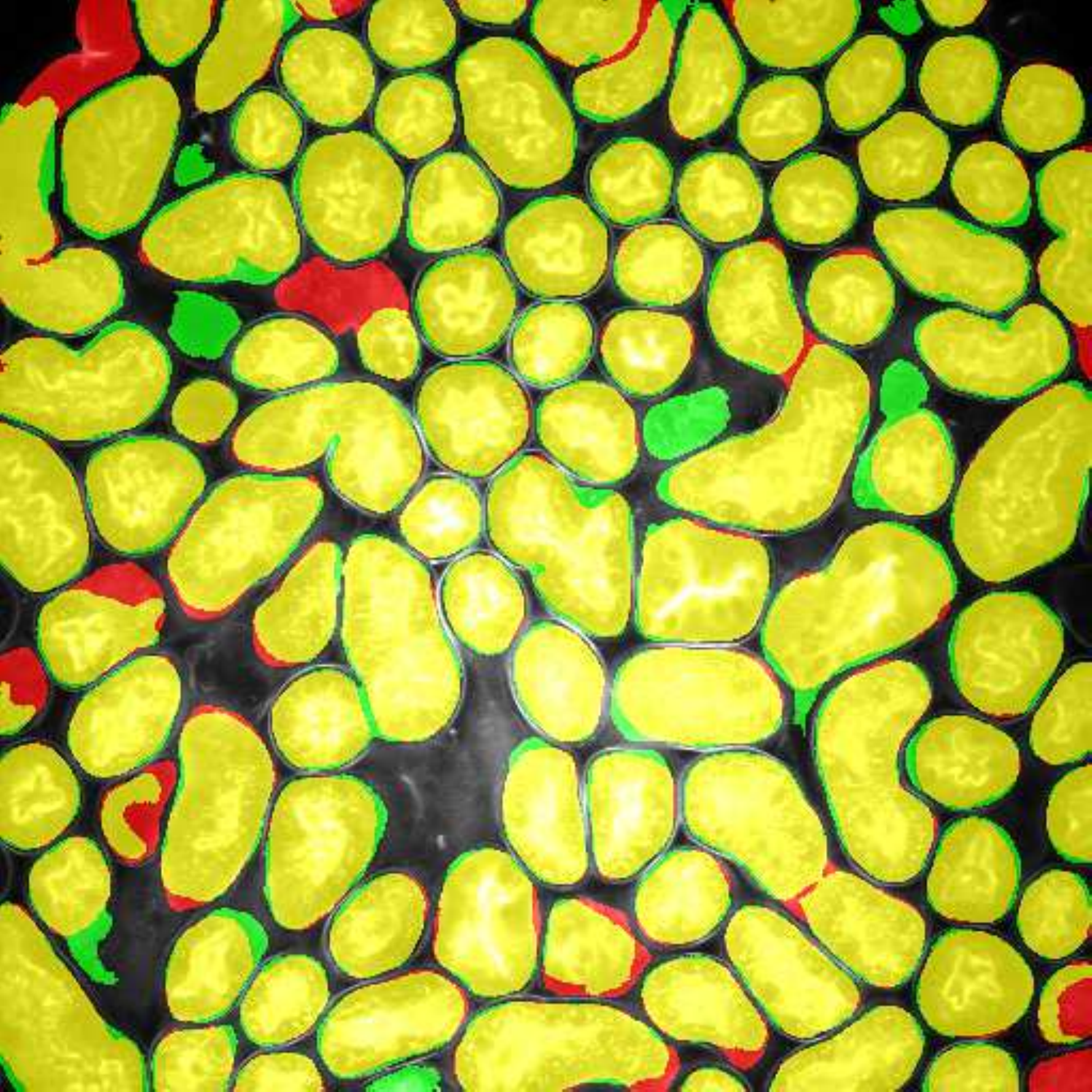}}\\
\caption{Qualitative evaluation/comparison of tubule segmentation results (shown in green) from the proposed method as well as other methods overlaid onto groundtruth image (shown in red) for $I_{z_{100}}$ belonging to  $Dataset-I$}
\label{fig:visComparison2} 
\end{figure}

The segmentation results shown in the second row generally missed many tubule regions. More specifically, \textit{3Dac} and \textit{3Dsquassh} could not capture the tubular structures but captured some in the center regions due to the intensity inhomogeneity of microscopy images. \textit{3DacIC} failed to segment  tubular structures but captured multiple lumens inside tubules as well as some tubule boundaries. In contrast, the segmentation results displayed in the third row showed falsely detected tubules. The main reason is that these tubule segmentation methods focused only on detecting boundaries of tubular structures. In particular, due to weak/blurry edges of fluorescence microscopy images, many boundaries were not continuous causing the filling operation to overflow from one tubule to another or to the background regions. The segmentation results using the CNN generally successfully segmented and identified each tubule region.

Figure \ref{fig:visComparison2} provides an alternative way to show the segmentation results. In particular, yellow regions correspond to true positives which are pixel locations that are identified as tubules in both the groundtruth and segmentation results. Green regions correspond to false positives which are pixel locations that are identified as background in groundtruth but tubules in segmentation results. Similarly, red pixels correspond to false negatives, namely pixel locations identified as tubules in the groundtruth but background in segmentation results, and black pixel regions correspond to true negative that are identified as background in both groundtruth and segmentation results. The green regions indicate Type-I error (false alarm) regions and the red regions represent Type-II error (miss) regions. As observed from Figure \ref{fig:visComparison2}, the segmentation results in the first row contained large red regions which mean large regions of tubules were missed. Conversely, the segmentation results shown in the second row contained many green regions indicating many background regions were falsely segmented as tubule regions. In contrast, the segmentation results in the third row had reasonably small green regions and red regions which indicate that the deep learning based segmentation results had higher pixel accuracy with relatively low Type-I and Type-II errors.

\subsection{Quantitative Evaluation}


In addition to the qualitative evaluation, quantitative metrics for evaluating the proposed method's segmentation accuracy of objects were utilized. In particular, we used pixel-based and object-based metrics. In the pixel-based metric, the pixel accuracy (PA), Type-I error, and Type-II error of pixel segmentation were obtained based on the manually annotated groundtruth images. Here, PA, Type-I, and Type-II are defined as below:
\begin{equation} \label{eq:PA}
PA = \frac{N^p_{tp} + N^p_{tn}}{N^p_{total}}, \quad Type-I = \frac{N^p_{fp}}{N^p_{total}}, \quad Type-II = \frac{N^p_{fn}}{N^p_{total}}
\end{equation}   
where $N^p_{tp}$, $N^p_{tn}$, $N^p_{fp}$, and $N^p_{fn}$ are defined to be  the number of segmented pixels that were labeled as true positives, true negatives, false positives, false negatives, respectively. $N^p_{total}$ denotes the total number of pixels in a image. These three pixel-based metrics obtained for $8$ different segmentation results are provided in Table \ref{tab:comp1}. As shown in Figure \ref{fig:visComparison2}, Type-II errors of the first three methods (\textit{3Dac}, \textit{3DacIC}, \textit{3Dsquassh}) were much higher compared to other methods. Similarly, Type-I errors of next three methods (\textit{Ellipse Fitting}, \textit{Jelly Filling}, \textit{Steerable Filter}) were much higher than those of the other methods. However, \textit{2DCNN} and \textit{2DCNNIC} had high PA and relatively low Type-I and Type-II errors.




In addition, our segmentation methods were evaluated using object-based criteria described in the \textit{2015 MICCAI Grand Segmentation Challenge} \cite{bib:DCAN, bib:Sirinukunwattana2017} namely: the F1 score metric, the Dice Index, and the Hausdorff Distance.

The F1 score metric is a measure of the segmentation/detection accuracy of individual objects. The evaluation of the F1 score metric is based on two metrics, precision $P$ and recall $R$. Denoting the number of tubules correctly identified by $N^o_{tp}$, the number of objects that are non-tubules but identified as tubules by $N^o_{fp}$, and the number of tubules that are not correctly identified as tubules by $N^o_{fn}$, respectively, then precision $P$ and recall $R$ are obtained as \cite{bib:DCAN}
\begin{equation}\label{eq:PR}
 P = \frac{N^o_{tp}}{N^o_{tp}+N^o_{fp}} \quad {\text {and}} \quad R = \frac{N^o_{tp}}{N^o_{tp}+N^o_{fn}}.
\end{equation} 
Given the values of $P$ and recall $R$, the F1 is found using
\begin{equation}\label{eq:F1}
 F1 = \frac{2PR}{P+R}.
\end{equation}
It is to be noted that a tubule segmented by the proposed method (or any other method for that matter) that overlaps at least $50\%$ with its corresponding manually annotated tubule is labeled as a true positive and added to the count of the true positives ($N^o_{tp}$), otherwise it is considered as a false positive and added to the count of the false positives ($N^o_{fp}$). Similarly, a manually annotated tubule that has no corresponding segmented tubule or overlaps less than $50\%$ with segmented tubular regions is considered to be a false negative and added to the count of the false negatives ($N^o_{fn}$). 


As mentioned above a second metric used to evaluate segmentation accuracy is the Dice Index (OD). The Dice Index \cite{bib:Dice1945} is a measure of similarity between two sets of samples. In our case, the two sets of samples are the sets of voxels belonging to a manually annotated tubule denoted by $G$, and the set of voxels belonging to a segmented tubule denoted by $S$. The Dice Index between $G$ and $S$ is defined as
\begin{equation}\label{eq:Dice}
D(G,S) = \frac{2 |G \cap S|}{|G|+|S|}
\end{equation}
where $|\cdot|$ denotes set cardinality which in this case will be the number of voxels belonging to an object. A higher value of the Dice Index indicates better segmentation match/results relative to the groundtruth data. A practical way of evaluating the Dice Index for segmented objects is described in \cite{bib:Sirinukunwattana2017} and is given by 
\begin{equation}\label{eq:ObjectDice}
D(G,S) = \frac{1}{2} \left [ \sum_{i=1}^{n_S} w_i D(G_i,S_i) + \sum_{j=1}^{n_G} \tilde{w}_j D(\tilde{G}_j,\tilde{S}_j) \right ]
\end{equation}
where
\begin{equation}\label{eq:Weights}
w_i = |S_i|/|\sum_{p=1}^{n_S} |S_p|, \quad \tilde{w}_j = |\tilde{G}_j|/|\sum_{q=1}^{n_G} |\tilde{G}_q|.
\end{equation}
In Eq (\ref{eq:ObjectDice}), $S_i$ denotes the $i^{th}$ tubule ($i \in \{1, \dots, n_S\}$) obtained by a segmentation method and $G_i$ denotes a manually annotated tubule that is maximally matched with $S_i$. Similarly, $\tilde{G}_j$ denotes the $j^{th}$ tubule ($j \in \{1, \dots, n_G\}$) identified in the groundtruth data and $\tilde{S}_j$ denotes a segmented tubule that is maximally matched with $\tilde{G}_j$. Finally, $n_S$ and $n_G$ denote the total number of segmented and manually annotated tubules, respectively. The first summation term in Eq (\ref{eq:ObjectDice}) represents how well each groundtruth tubule overlaps with its segmented counterpart, whereas the second summation term represents how well each segmented tubule overlaps with its manually annotated counterpart. The terms $w_i$ and $\tilde{w}_j$ which are used to weight the summation terms represent the fraction of the space that each tubule $S_i$ and $\tilde{G}_j$ occupies within the entire tubule region, respectively. 


While the Dice Index measures segmentation accuracy, a third metric, the Hausdorff Distance (OH), is needed to evaluate shape similarity. The Hausdorff Distance \cite{bib:Huttenlocher1993}, $H(G,S)$, between a segmented tubule $S$ and its manually annotated counterpart $G$, is defined to be 
\begin{equation}\label{eq:Hausdorff}
H(G,S) = \max \{ \underset{x \in G} \sup \,\,  \underset{y \in S} \inf ||x- y||_2, \,\, \underset{y \in S} \sup \,\,  \underset{x \in G} \inf ||x- y||_2 \}.
\end{equation}
Here, $||x - y||_2$ denotes the Euclidean distance between a pair of pixels $x$ and $y$. Based on Eq (\ref{eq:Hausdorff}), the Hausdorff Distance obtains the maximum distance among all pairs of voxels on the boundaries of $S$ and $G$. Therefore, a smaller value of the Hausdorff Distance indicates a higher similarity in shape between the boundaries of $S$ and $G$. As done above (see Eq (\ref{eq:ObjectDice})), a practical way of finding the Hausdorff Distance between a segmented tubule $S$ and its manually annotated counterpart $G$ is given by \cite{bib:Sirinukunwattana2017}:
\begin{equation}\label{eq:ObjectHausdorff}
H(G,S) = \frac{1}{2} \left [ \sum_{i=1}^{n_S} w_i H(G_i,S_i) + \sum_{j=1}^{n_G} \tilde{w}_j H(\tilde{G}_j,\tilde{S}_j) \right ]
\end{equation}  
where the parameters $w_i$ and $\tilde{w}_j$ are defined in Eq (\ref{eq:Weights}).

\begin{figure}[ht!]
\vspace{-0.1in}
	\centering
	\subfloat[$I^O_{z_{100}}$ of $Dataset-I$]
		 {\label{fig:origDataI100}\includegraphics[width=0.15\textwidth]{water-scale-mount_red_z0100}}
	\,\,\,
	\subfloat[$I^O_{z_{150}}$ of $Dataset-I$]
		 {\label{fig:origDataI150}\includegraphics[width=0.15\textwidth]{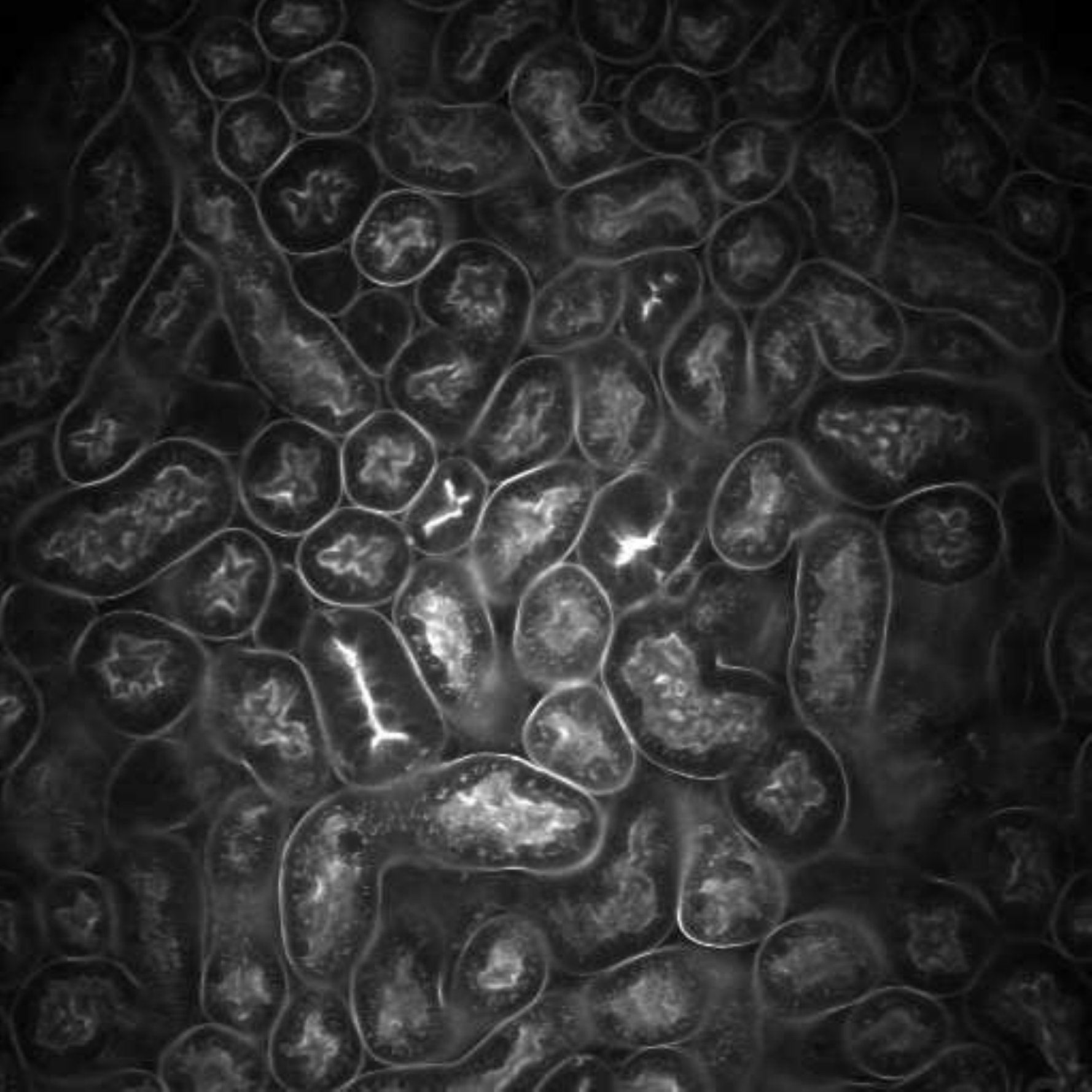}}
	\,\,\,
 	\subfloat[$I^O_{z_{200}}$ of $Dataset-I$]
		 {\label{fig:origDataI200}\includegraphics[width=0.15\textwidth]{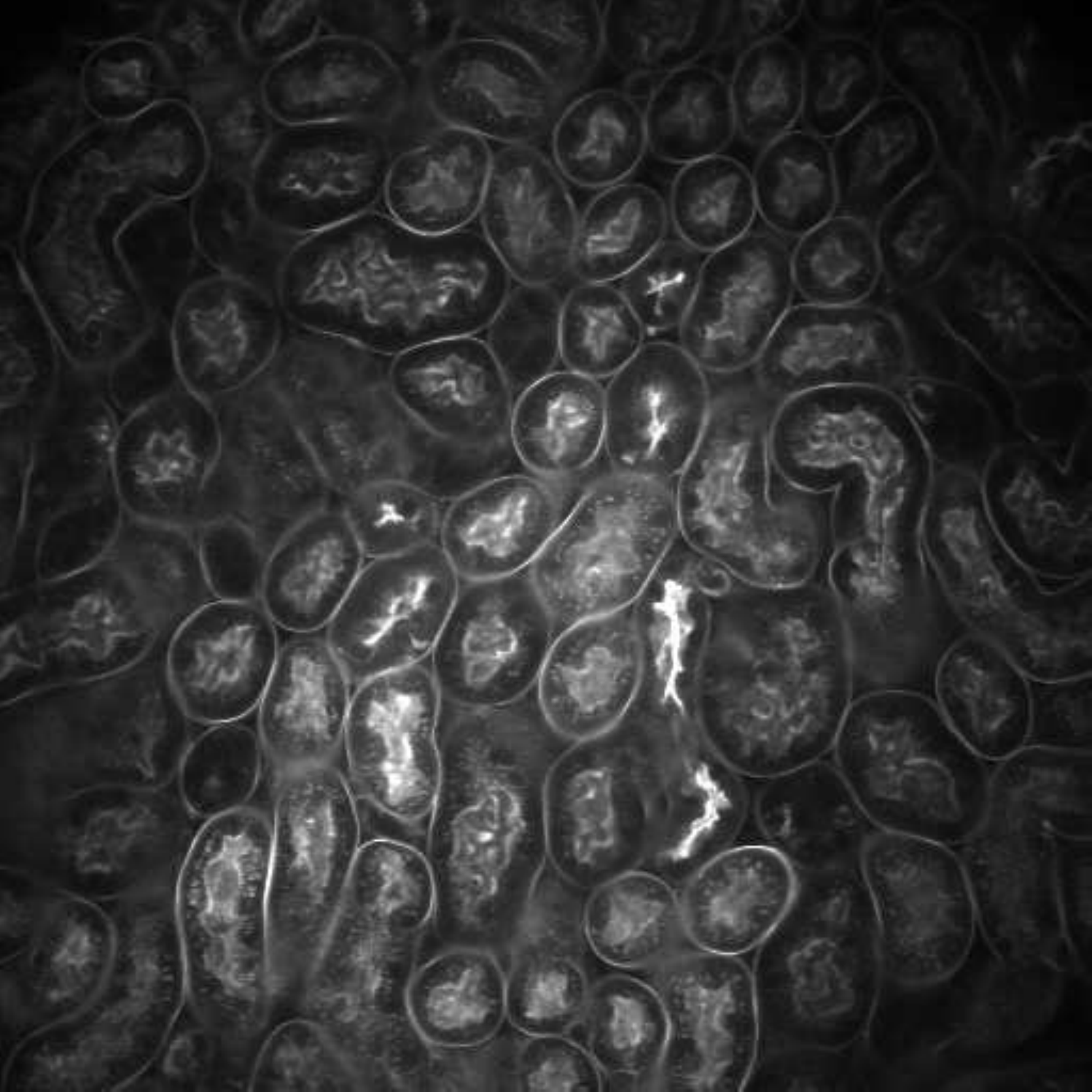}}\\
\vspace{-0.15in}
	\,\,\,
	\subfloat[$I^F_{z_{100}}$ of $Dataset-I$]
		 {\label{fig:segOverlaidRGBDataI100}\includegraphics[width=0.15\textwidth]{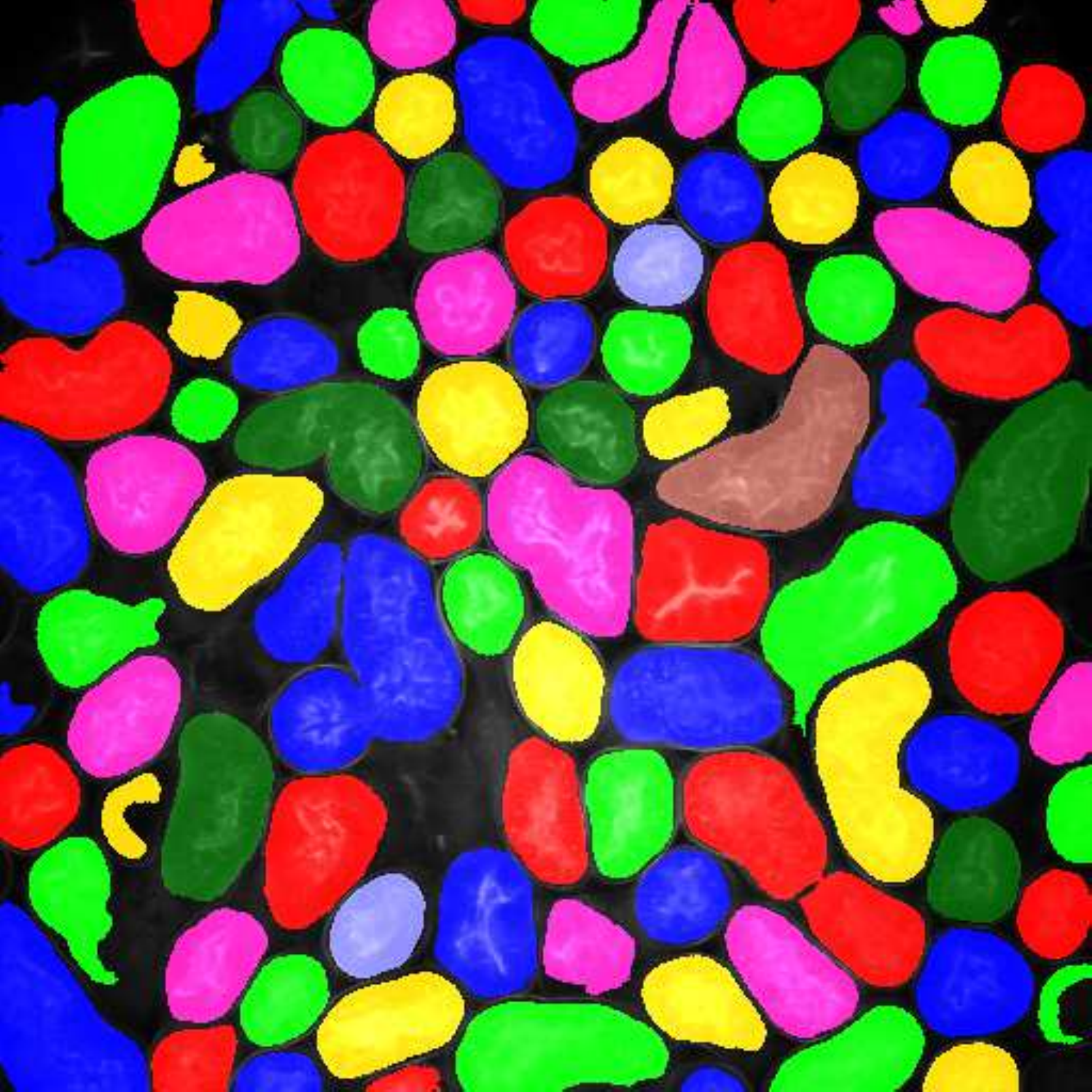}}
	\,\,\,
	\subfloat[$I^F_{z_{150}}$ of $Dataset-I$]
		 {\label{fig:segOverlaidRGBDataI150}\includegraphics[width=0.15\textwidth]{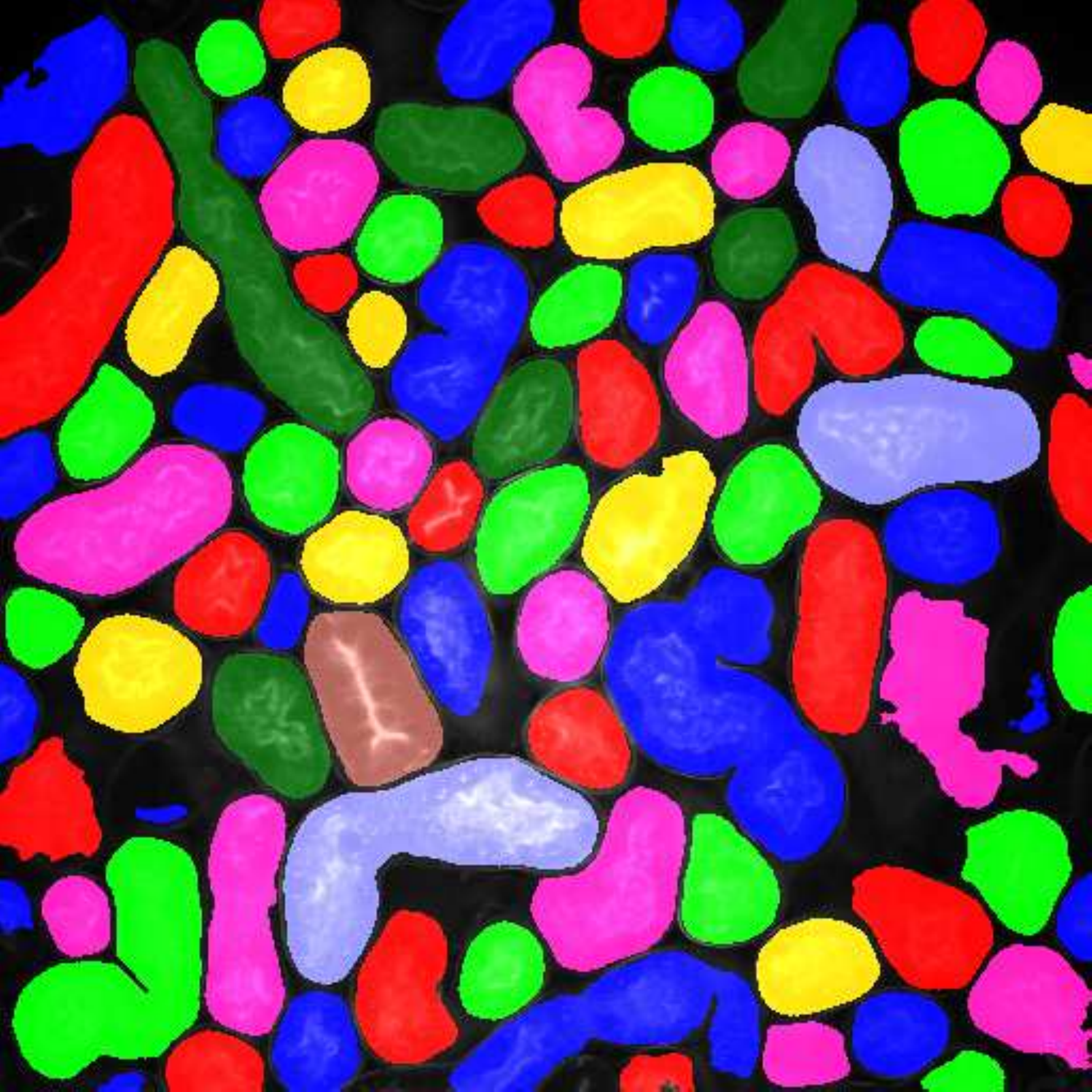}} 	
	\,\,\,
	\subfloat[$I^F_{z_{200}}$ of $Dataset-I$]
		 {\label{fig:segOverlaidRGBDataI200}\includegraphics[width=0.15\textwidth]{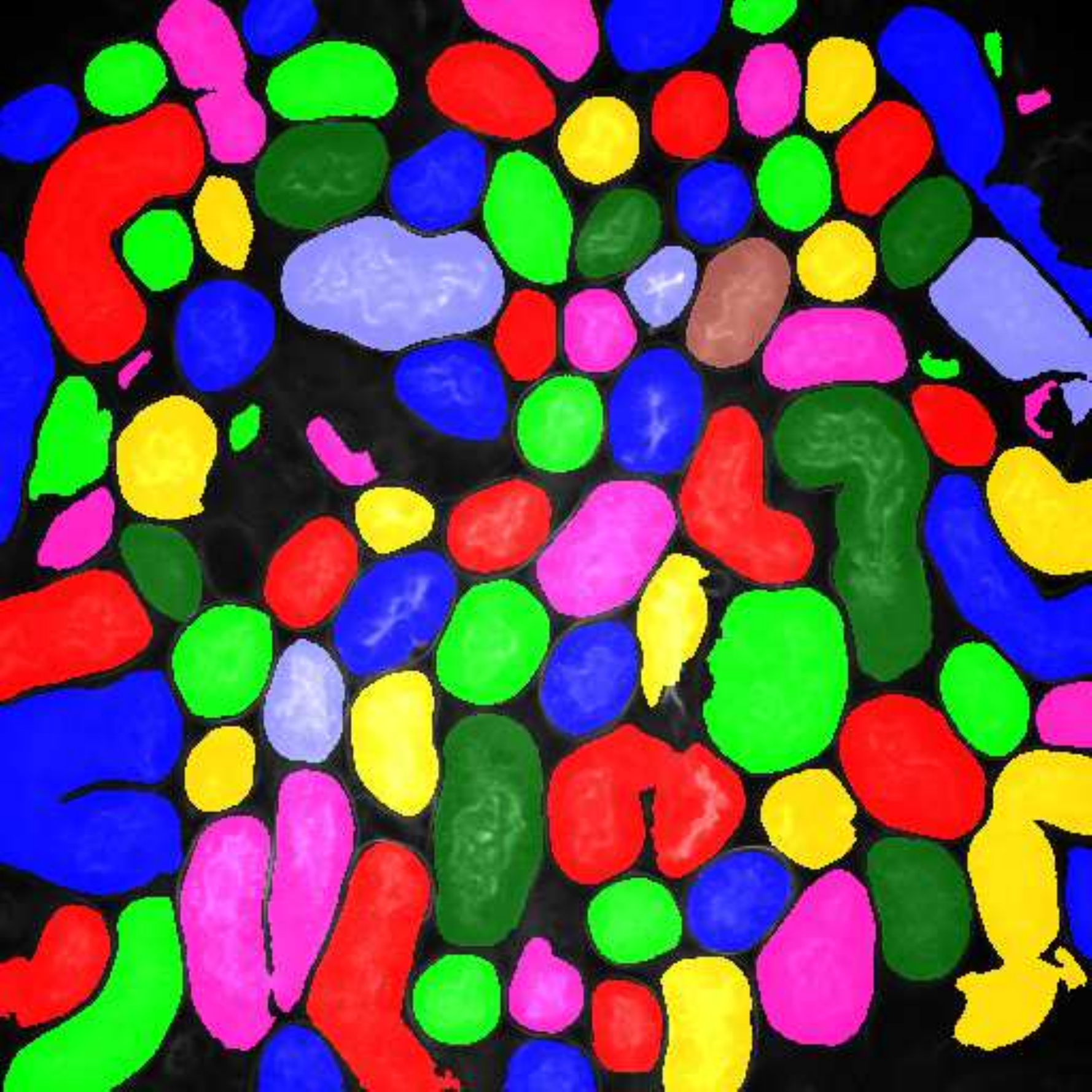}} \\
\vspace{-0.15in}
	\,\,\,
	\subfloat[$I^O_{z_{50}}$ of $Dataset-II$]	
		 {\label{fig:origDataII50}\includegraphics[width=0.15\textwidth]{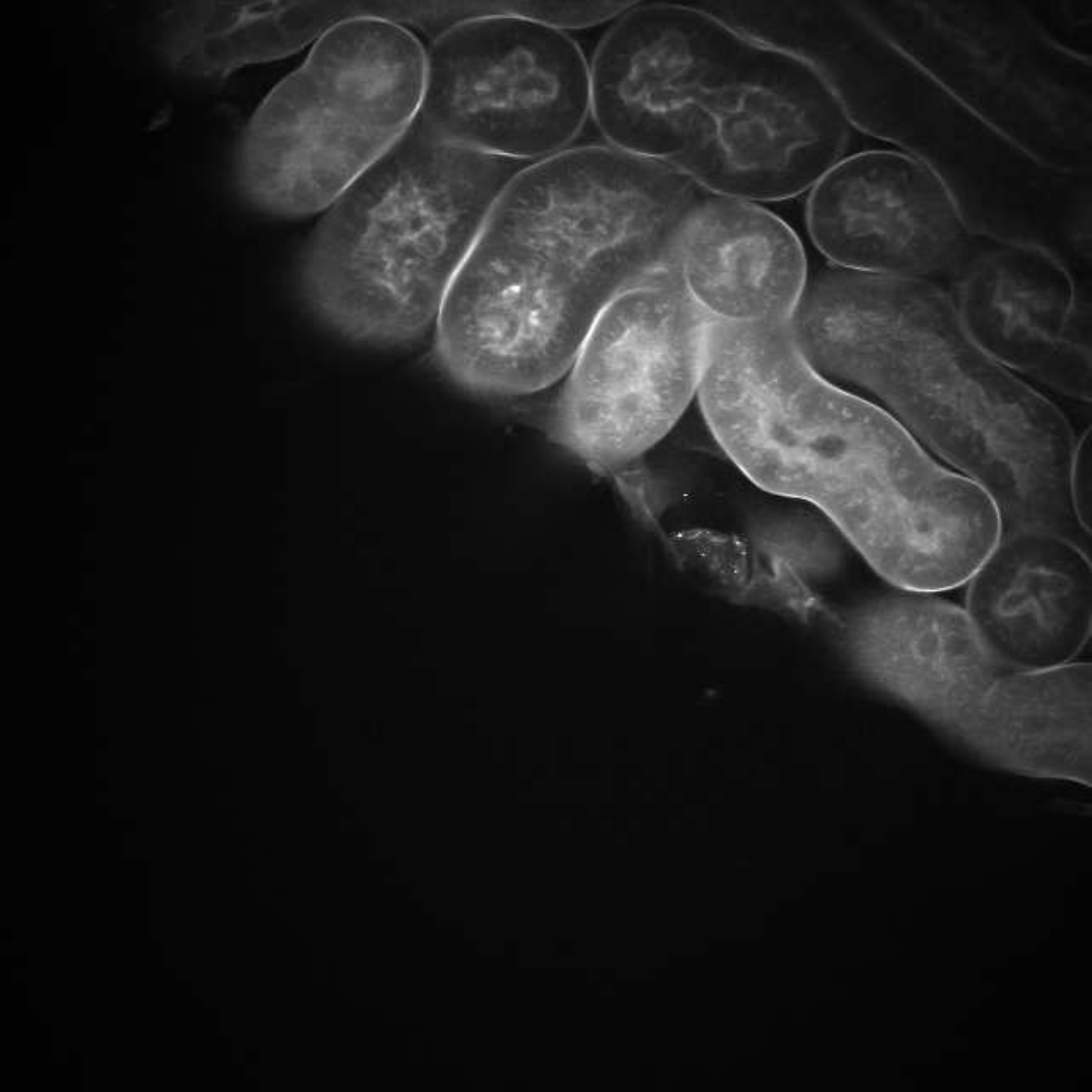}}
		 \,\,\,
	\subfloat[$I^O_{z_{150}}$ of $Dataset-II$]
		 {\label{fig:origDataII150}\includegraphics[width=0.15\textwidth]{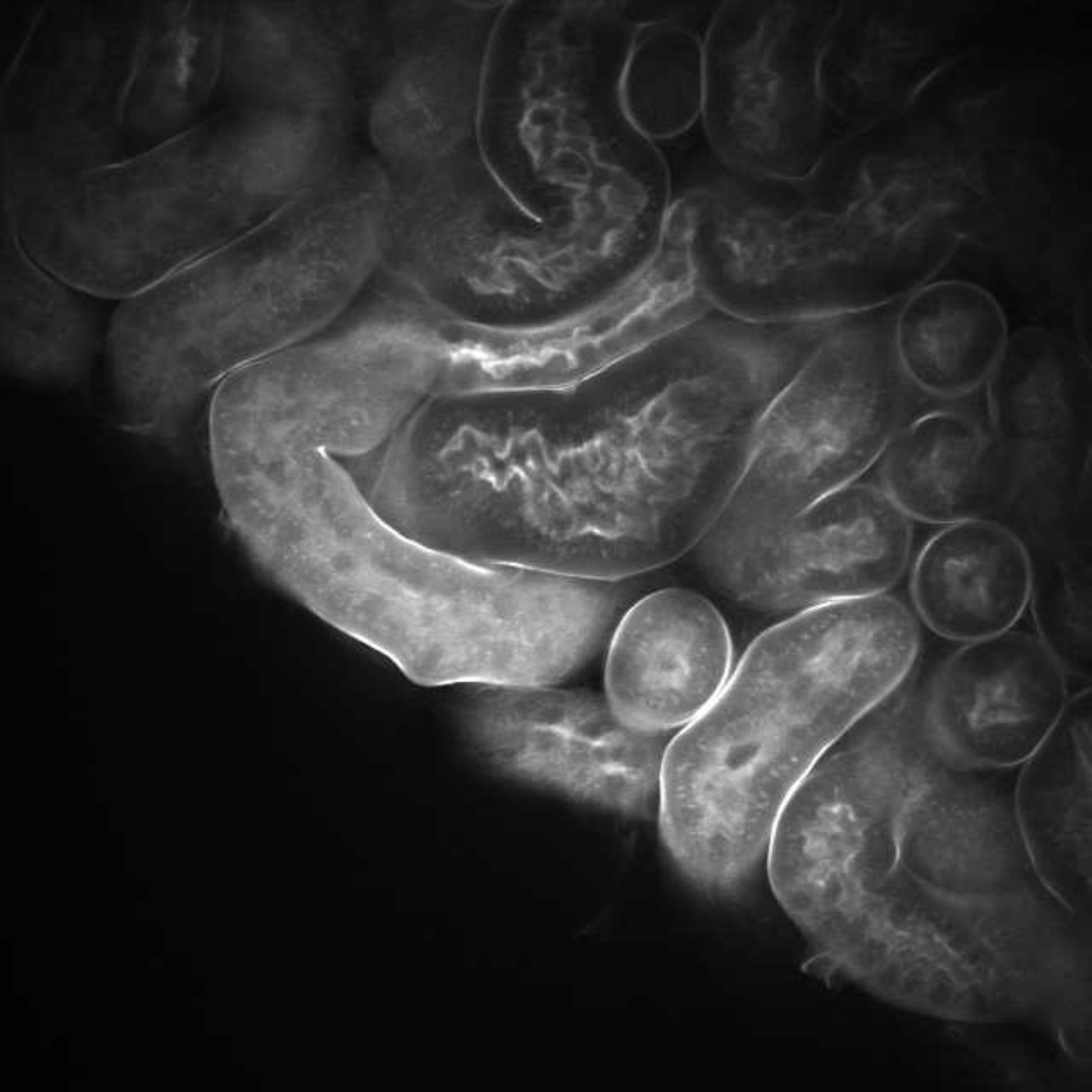}}
		 \,\,\,
	\subfloat[$I^O_{z_{250}}$ of $Dataset-II$]
		 {\label{fig:origDataII250}\includegraphics[width=0.15\textwidth]{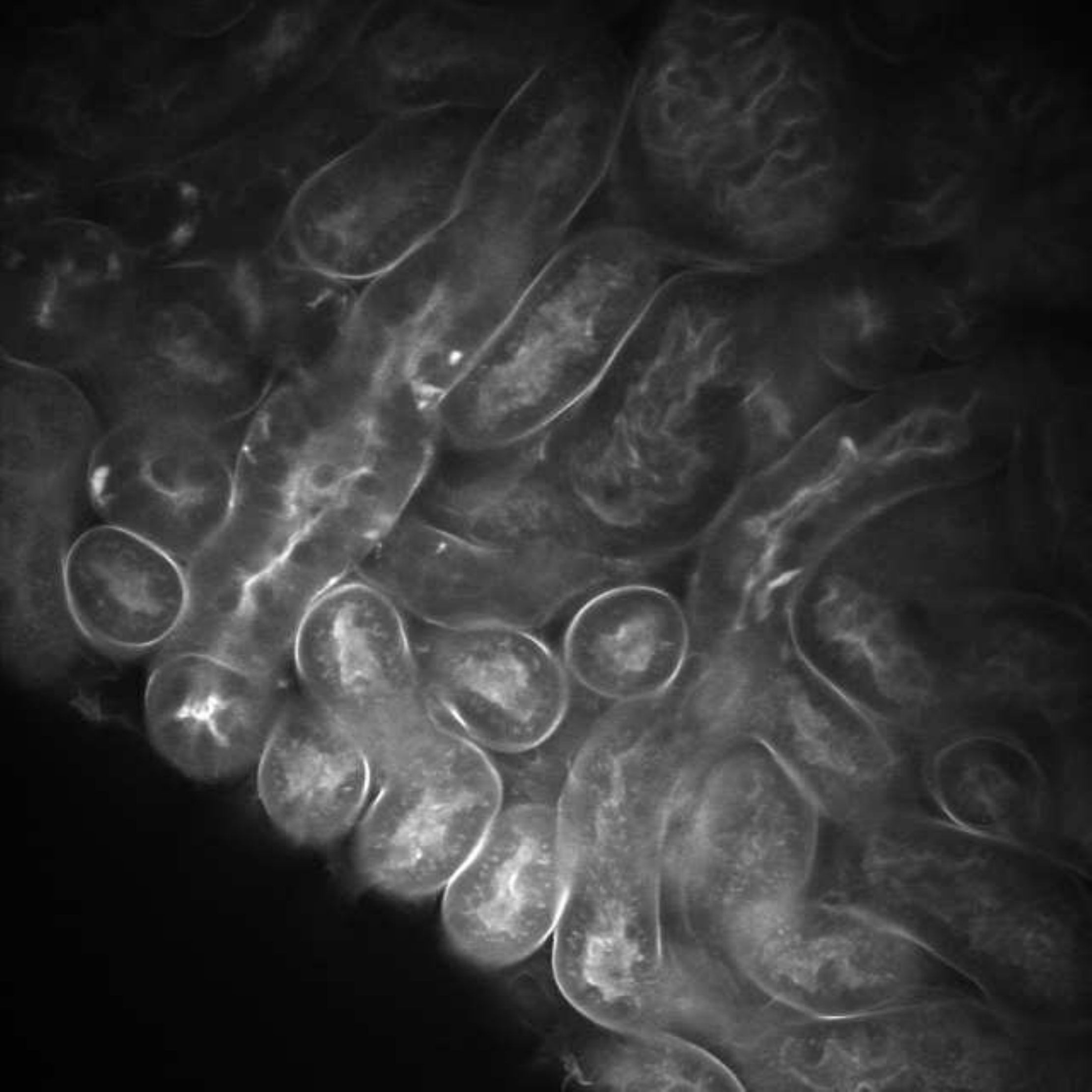}}\\
\vspace{-0.15in}
	\,\,\,
	\subfloat[$I^F_{z_{50}}$ of $Dataset-II$]
		 {\label{fig:segOverlaidRGBDataII50}\includegraphics[width=0.15\textwidth]{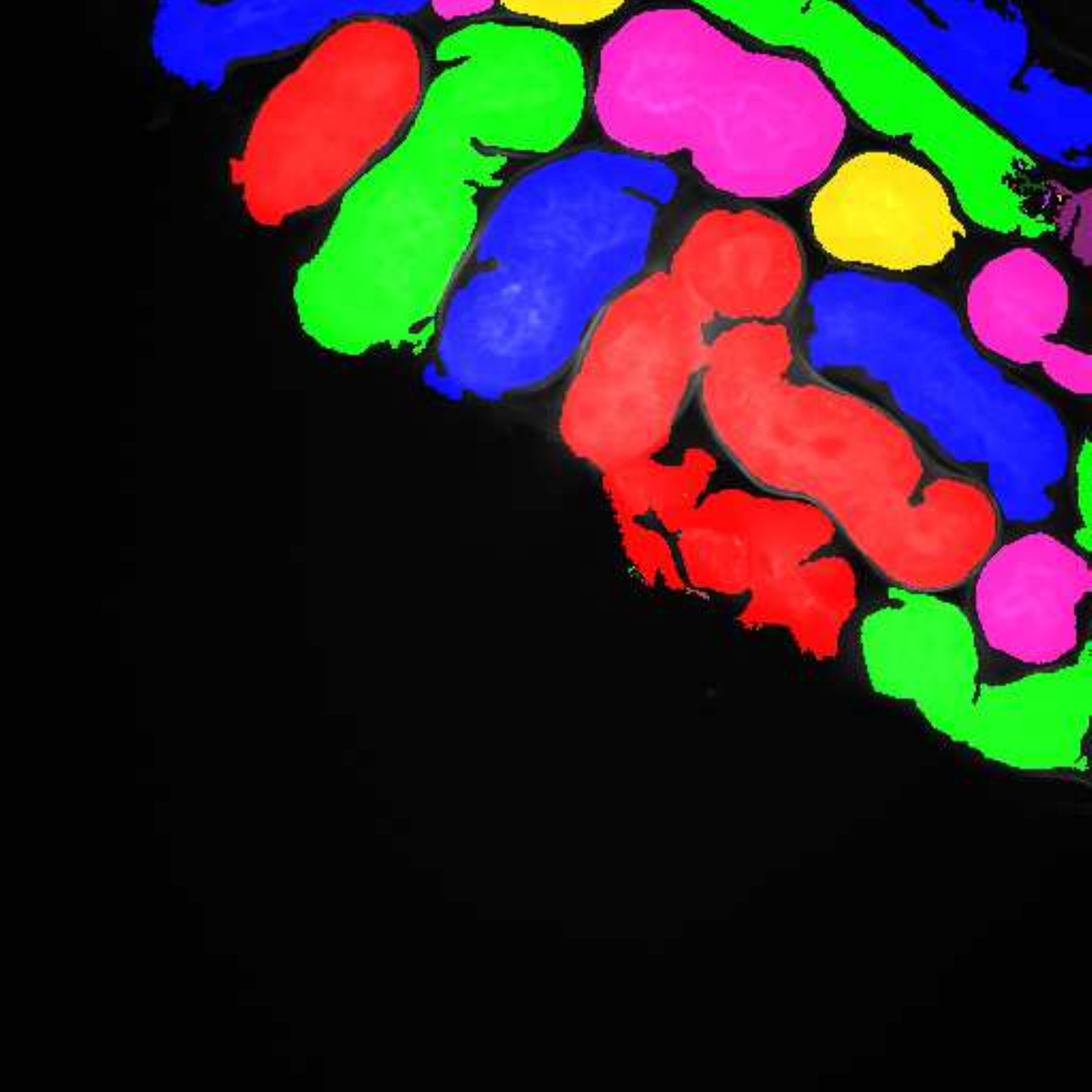}}
	\,\,\,
	\subfloat[$I^F_{z_{150}}$ of $Dataset-II$]
		 {\label{fig:segOverlaidRGBDataII150}\includegraphics[width=0.15\textwidth]{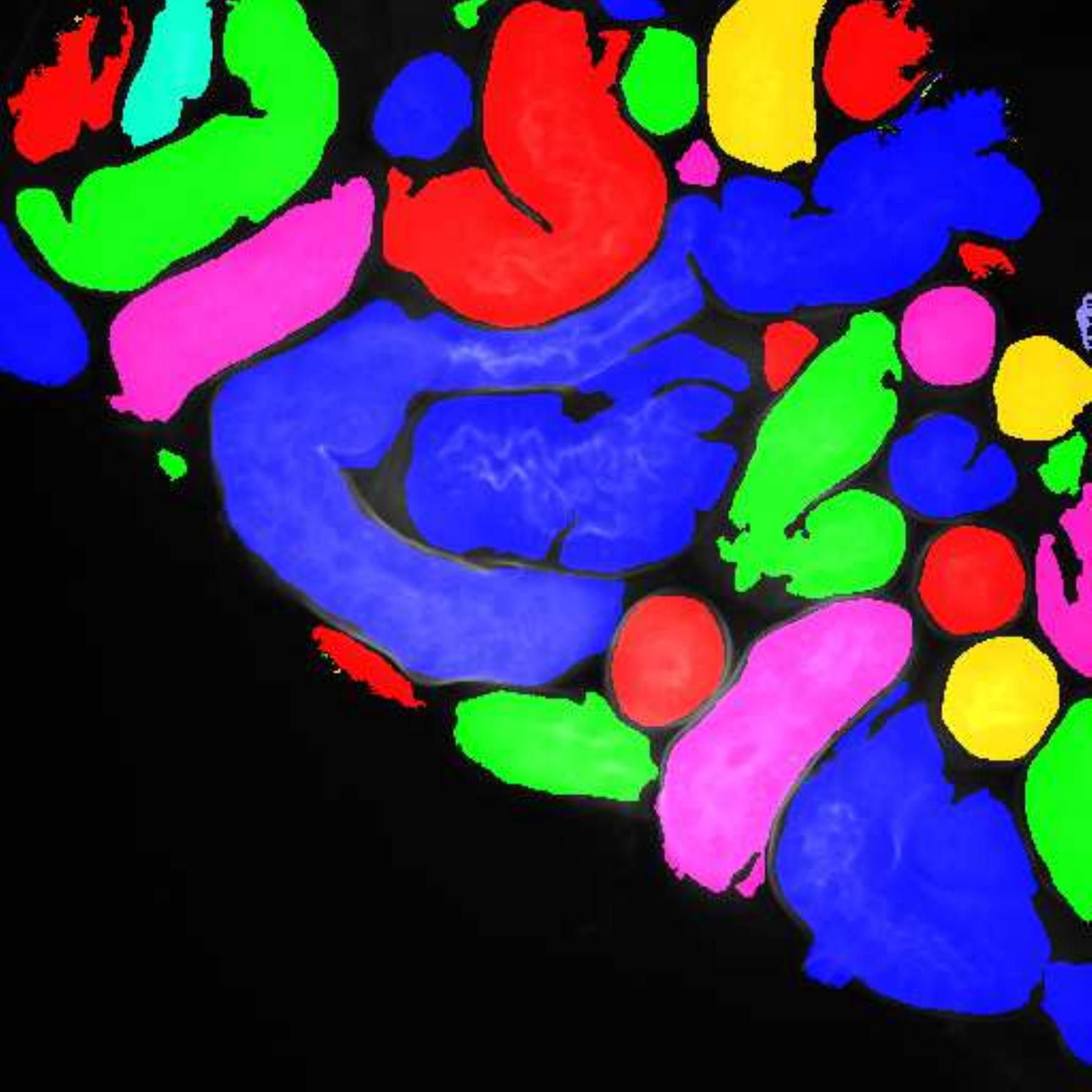}}
	\,\,\,
 	\subfloat[$I^F_{z_{250}}$ of $Dataset-II$]
		 {\label{fig:segOverlaidRGBDataII250}\includegraphics[width=0.15\textwidth]{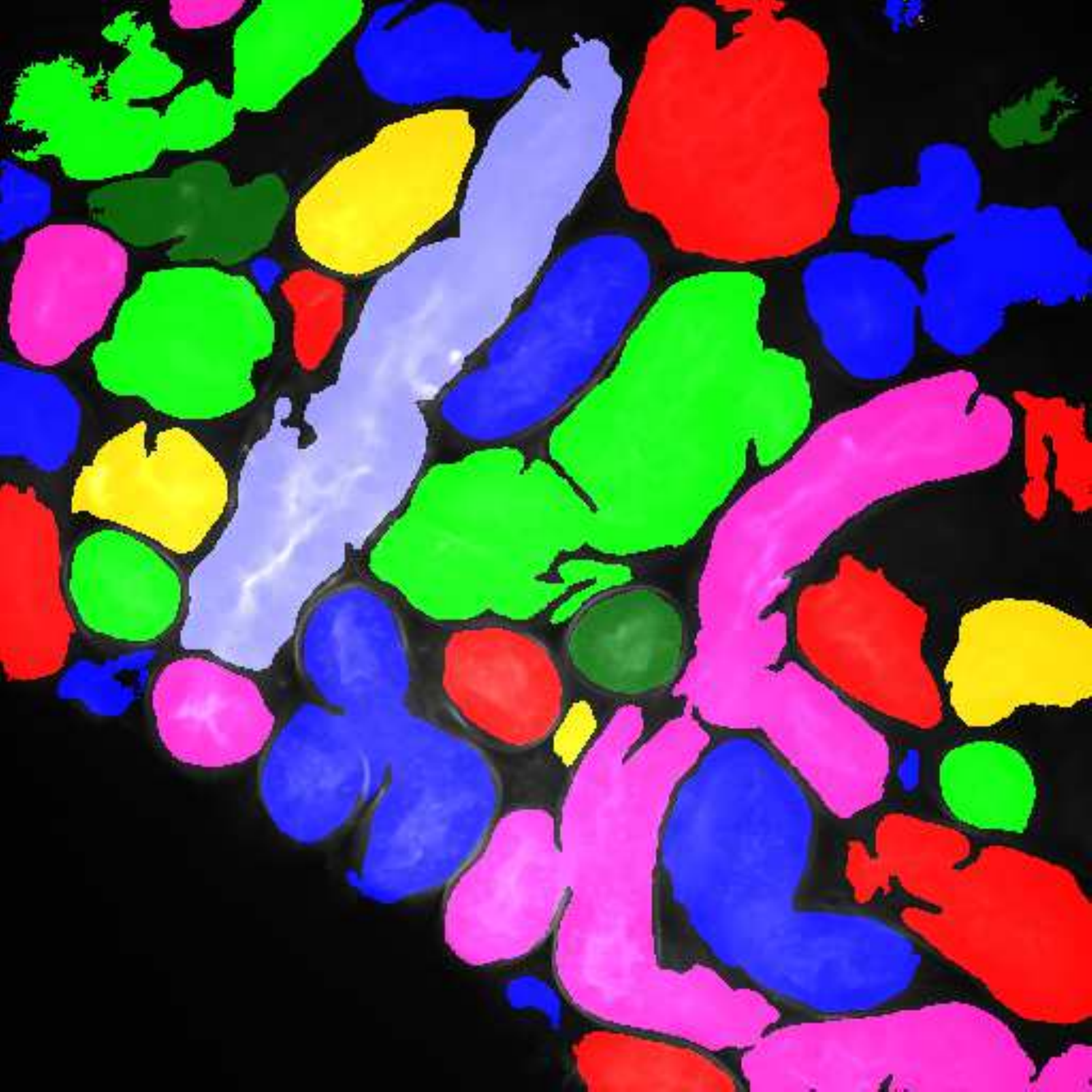}}
		 \\
\caption{Original and color coded segmentation results of the proposed method on different depth of $Dataset-I$ and $Dataset-II$ using same trained model $\mathcal{M}$}
\label{fig:visColorCoded} 
\end{figure}

The performance of the proposed method and other methods based on the F1 score, OD, and OH metrics were obtained and tabulated in Table \ref{tab:comp1}. As mentioned above higher values of F1 and OD are considered to be indicators of better segmentation results. In contrast, lower values of OH indicate better segmentation result. As can be seen in Table \ref{tab:comp1}, our proposed method outperformed all the other segmentation methods against which the proposed method is being evaluated. In particular, \textit{3Dac}, \textit{3DacIC}, \textit{3Dsquassh}, and \textit{Steerable Filter} had low F1 scores since these segmentation methods had large Type-I or Type-II errors.
Similarly, all of the methods except for \textit{2DCNN} suffered from low OD and high OH values. In particular, since the segmentation results of \textit{3Dsquassh}, \textit{Ellipse Fitting}, and \textit{Steerable Filter} failed to distinguish most of the individual tubules, they exhibited low OD and high OH values. Note that \textit{3DacIC} had relatively low OH {\em and} low OD values since it segmented some tubule boundaries as well as some partial regions (lumen) inside the tubules. Lastly, the use of intensity inhomogeneity correction in the proposed method improved its performance relative to that of \textit{2DCNN}.

For visual evaluation we provide the segmentation results of the proposed method using two different datasets: $Dataset-I$ and $Dataset-II$, sampled at different depths within the volumes. The first row shows original microscopy images $I^O_{z_{100}}$, $I^O_{z_{150}}$, and $I^O_{z_{200}}$ from $Dataset-I$ and the second row displays the segmentation results corresponding to the first row. To better visualize the segmentation results, we highlighted individual tubules with different colors and overlaid them onto the original microscopy images. Similarly, the third row exhibits original microscopy images $I^O_{z_{50}}$, $I^O_{z_{150}}$, and $I^O_{z_{250}}$ from $Dataset-II$. Their corresponding segmentation results are shown in the fourth row. Note that the model $\mathcal{M}$ which was trained on $Dataset-I$ was used for $Dataset-II$ during the inference stage. Although the shape, size, and orientation of tubular structures presented in $Dataset-II$ are all different from $Dataset-I$, the proposed method can still successfully segment and identify individual tubules presented in $Dataset-II$ as well as individual tubules in $Dataset-I$.

\section{Conclusions} \label{sec:con}
This paper presented a tubular structure segmentation method that uses inhomogeneity correction, data augmentation, a convolutional neural network, and postprocessing.
The qualitative and quantitative results indicate that the proposed method can successfully segment and identify individual tubules compared to other segmentation methods. In the future, we plan to utilize 3D information generated from realistic 3D synthetic tubular structures to improve segmentation results as well as reduce manual annotation work.

\section{Acknowledgment}
This work was partially supported by a George M. O'Brien Award from the National Institutes of Health NIH/NIDDK P30 DK079312.

\small
\bibliographystyle{IEEEbib}   
\bibliography{EI2018_bib}

\begin{biography}
Soonam Lee received the B.S. (Summa Cum Laude) degree in Electrical and Computer Engineering from Hanyang University, Seoul, Korea in 2008. He received the M.S. degree in Electrical Engineering and Computer Science from University of Michigan, MI in 2012. He also received the M.S. degree in Mathematics from University of Michigan, MI in 2012. He is currently pursuing Ph.D. at the School of Electrical and Computer Engineering, Purdue University, West Lafayette, IN. His research interests include image processing, computer vision, machine learning, deep learning and biomedical imaging area.

Chichen Fu received his B.S. in Electrical Engineering from Purdue University in 2014. He is currently  pursuing his Ph.D. in Electrical and Computer Engineering from Purdue University. His research interests lie in machine learning, image processing and computer vision.

Paul Salama is Professor of Electrical and Computer Engineering and Assistant Dean of Graduate Programs at the Purdue School of Engineering and Technology at the Indiana University Purdue University, Indianapolis. His research interests include image and video processing, image analysis, image and video compression, medical imaging, and machine learning. Dr. Salama is a Senior Member of the IEEE and member of the SPIE.

Kenneth W. Dunn was born in Redwood City, California. He received his Ph.D. at the State University of New York at Stony Brook, and completed his post-doctoral training at Columbia University, where he was subsequently appointed as an Assistant Professor. In 1995 he took a position in the Nephrology Division in the School of Medicine at Indiana University, where he subsequently rose to the rank of Professor of Medicine and Biochemistry. His research interests include epithelial cell biology, drug-induced liver injury and quantitative light microscopy. In his role as Director of the Indiana Center for Biological Microscopy, he has overseen the development quantitative and intravital microscopy at the IU School of Medicine, directing core services for NIH program projects in Hematology, Diabetes and Nephrology.

Edward J. Delp was born in Cincinnati, Ohio. He is currently The Charles William Harrison Distinguished Professor of Electrical and Computer Engineering and Professor of Biomedical Engineering at Purdue University. His research interests include image and video processing, image analysis, computer vision, image and video compression, multimedia security, medical imaging, multimedia systems, communication and information theory. Dr. Delp is a Life Fellow of the IEEE, a Fellow of the SPIE, a Fellow of IS\&T, and a Fellow of the American Institute of Medical and Biological Engineering.

\end{biography}

\end{document}